\documentclass{article}

\usepackage{microtype}
\usepackage{graphicx}
\usepackage{subcaption}
\usepackage{booktabs} 

\usepackage{hyperref}

\usepackage[accepted]{icml2026}

\usepackage{amsmath}
\usepackage{amssymb}
\usepackage{mathtools}
\usepackage{amsthm}
\usepackage{longtable}
\usepackage{booktabs} 
\usepackage{multirow}

\usepackage[capitalize,noabbrev]{cleveref}

\theoremstyle{plain}

\theoremstyle{definition}

\theoremstyle{remark}

\usepackage[textsize=tiny]{todonotes}

\icmltitlerunning{ForensicConcept: Transferable Forensic Concepts for AIGI Detection}

\begin{document}

\twocolumn[
  \icmltitle{\textsc{ForensicConcept}: Transferable Forensic Concepts for AIGI Detection}



  \icmlsetsymbol{equal}{*}

  \begin{icmlauthorlist}
    \icmlauthor{Menyanshu Zhou}{equal,yyy}
    \icmlauthor{Ziyin Zhou}{equal,yyy}
    \icmlauthor{Ke Sun}{yyy}
    \icmlauthor{Yunpeng Luo}{yyy}
    \icmlauthor{Jiayi Ji}{yyy}
    \icmlauthor{Xiaoshuai Sun}{yyy,comp}
    \icmlauthor{Rongrong Ji}{yyy}
  \end{icmlauthorlist}

  \icmlaffiliation{yyy}{Key Laboratory of Multimedia Trusted Perception and Efficient Computing, Ministry of Education of China, School of
Informatics, Xiamen University, Xiamen 361005, P.R.China}
  \icmlaffiliation{comp}{Sino-Russian Research Center for Digital Economy}

  \icmlcorrespondingauthor{Xiaoshuai Sun}{xssun@xmu.edu.cn}

  \icmlkeywords{Machine Learning, ICML}

  \vskip 0.3in
]



\printAffiliationsAndNotice{\icmlEqualContribution}  
\begin{abstract}
AI-generated image detectors achieve high accuracy on in-distribution data but often fail on unseen generators. A key obstacle to understanding this failure is the black-box nature of current detectors: they do not reveal which evidence drives their decisions. We propose \textsc{ForensicConcept}, a framework that extracts explicit forensic concepts from detectors and enables their transfer across backbones. Our method localizes decision-critical patches via Transformer attribution, clusters them into a compact concept codebook, and uses a concept-aligned projection to produce auditable evidence readouts. Motivated by prior studies showing that DINO representations can guide diffusion generation and exhibit concept-level correspondence with diffusion features, we introduce a generation-trace reference based on CleanDIFT diffusion features and quantify backbone-trace alignment via neighborhood-structure consistency (CKNNA). We further propose concept codebook injection to transfer diffusion-derived concepts into target backbones. Experiments on GenImage, GAN-family, and Chameleon benchmarks show consistent improvements over prior methods. We also find that CKNNA alignment predicts transfer effectiveness, providing a principled explanation for why some backbones yield more transferable forensic evidence than others. Code is available at \url{https://github.com/EthanAdamm/FORENSICCONCEPT}.
\end{abstract}

\begin{figure}[!t]
\begin{center}
\centerline{\includegraphics[width=\columnwidth]{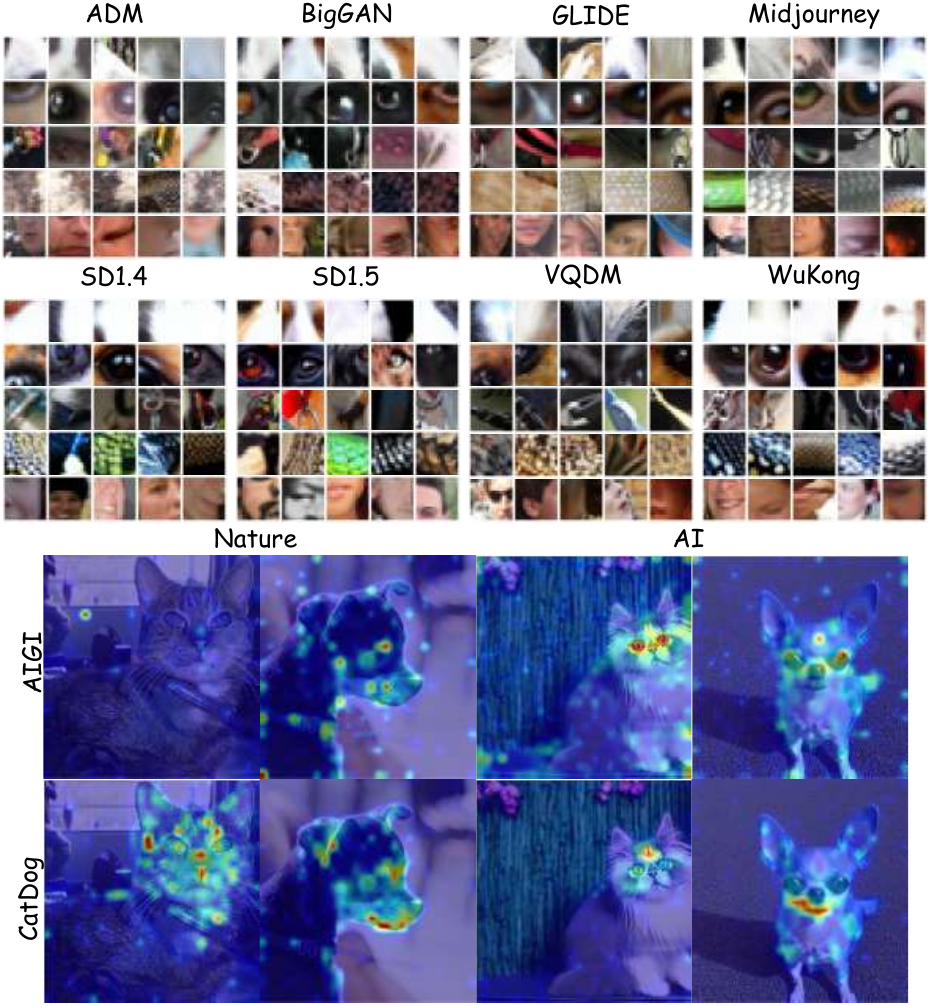}}
\caption{\textbf{Forensic evidence is diffuse yet conceptualizable.}
\textbf{Top:} Dataset-specific forensic codebooks (DINOv3 with injected AIGI prior); patches nearest to concept centers reveal recurring low-level patterns across generators.
\textbf{Bottom:} Attribution maps show a mismatch: an AIGI detector attends to diffuse cues, while a semantic classifier (cat vs.\ dog) focuses on object parts.}

\label{fig1}
\end{center}
\vskip -0.2in
\end{figure}
\section{Introduction}
AI-generated image detection aims to distinguish synthetic images from real photographs. This task has become increasingly critical as generative models advance rapidly \cite{goodfellow2014generative,karras2020analyzing,ho2020denoising,rombach2022high}. GANs and diffusion models can now synthesize photorealistic images at scale, enabling malicious applications such as misinformation and fraud. Reliable detection methods are essential for content authenticity and digital forensics.

Most existing detectors formulate this task as binary classification, training networks to produce a single ``fake likelihood'' score \cite{wang2020cnn,zhu2023genimage}. These methods achieve high accuracy on in-distribution data but often fail on unseen generators. A natural question arises: \textit{why do detectors fail to generalize?} One hypothesis is that they learn generator-specific shortcuts rather than transferable forensic traces. However, verifying this hypothesis is difficult because current detectors are black boxes. They do not reveal what evidence drives their decisions. Without understanding the underlying evidence, we cannot diagnose generalization failures or design principled solutions.

Recent work has explored multiple directions to improve generalization. Some methods scale up training data to cover more generators \cite{zhu2023genimage,schinas2024sidbench}. Others leverage pretrained representations from vision-language models \cite{ojha2023towards,cozzolino2024raising}. Alternative signals such as reconstruction error and frequency patterns have also been explored \cite{wang2023dire,luo2024lare}. While these approaches improve empirical performance, they still treat the detector as a black-box. The question of what evidence is learned and whether it transfers remains unanswered.

We take a different approach by directly examining the evidence that detectors rely on. In \cref{fig1}, we visualize and compare the attribution patterns of an AI-generated image detector with a semantic classifier. The semantic classifier focuses on object parts like eyes and ears to distinguish cats from dogs. In contrast, the forensic detector attends to scattered regions across the image, including backgrounds, textures, and smooth areas. This confirms that forensic cues are spatially diffuse and fundamentally different from semantic cues. A key question is whether such diffuse evidence can be explicitly characterized and transferred.

We further investigate the structure of this diffuse evidence. By clustering the patches that detectors attend to, we find that they form coherent groups with repeatable low-level patterns. The top panel of \cref{fig1} shows examples: patches within each cluster share similar textures or edge statistics, and these patterns recur across images from different generators. This observation suggests that forensic evidence, though spatially scattered, exhibits structured geometry in feature space. We call these recurring patterns \emph{forensic concepts} and ask: \textit{can we extract such concepts explicitly and transfer them across detectors?}
Crucially, do these concepts actually capture genuine generator traces, or merely detector-specific shortcuts?

Based on this insight, we propose \textsc{ForensicConcept}, a framework that converts diffuse forensic evidence into explicit, transferable units. Our method has three components. First, we perform forensic concept induction. We train a detector with adapter-guided tuning, localize decision-critical patches via Transformer attribution, and cluster the corresponding tokens to form a compact codebook. A concept-aligned projection maps representations into this space, producing auditable evidence readouts alongside predictions. Second, we introduce a diffusion-trace reference as an external, generator-grounded space for validating and comparing forensic evidence across backbones, motivated by the observed correspondence between DINO \cite{oquab2023dinov2} and diffusion representations \cite{zheng2026diffusion,zhang2025both,yu2025representation}. We instantiate this reference with CleanDIFT features \cite{stracke2025cleandift}, using them as an external generation-trace reference rather than as an oracle. This allows us to assess whether detector evidence preserves neighborhood structures consistent with diffusion traces. We quantify the geometric alignment between the detector’s evidence and these diffusion traces using neighborhood-structure consistency (CKNNA) \cite{huh2024position}. Third, we perform concept codebook injection. We construct diffusion-derived codebooks and inject them into target backbones, enabling cross-generator transfer with visualizable prototypes.

We validate our framework on three benchmarks. On GenImage \cite{zhu2023genimage}, our method achieves 92.0\% mean accuracy across the official test subsets, outperforming prior detectors. On the GAN-family benchmark \cite{zhong2023patchcraft}, training on diffusion data and transferring to GAN generators yields 90.1\% accuracy. On Chameleon \cite{yan2025sanity} with stronger distribution shifts, we reach 84.4\% accuracy. We also find that CKNNA alignment predicts transfer effectiveness: backbones with stronger alignment to diffusion traces yield more transferable concepts.

Our contributions are:
\begin{itemize}
    \item We introduce forensic concepts into AI-generated image detection by converting attribution-localized evidence patches into reusable units with auditable evidence readouts.
    \item We propose CKNNA as a quantitative metric to measure alignment between backbone evidence and generation-trace references, explaining transferability differences across backbones.
    \item We demonstrate consistent cross-generator improvements via concept codebook injection, with visualizable prototypes that make forensic evidence inspectable.
\end{itemize}

\section{Related work}
\subsection{AI-Generated Image Detection and Generalization}
AI-generated image (AIGI) detection has been extensively studied under both CNN and ViT-style backbones, with a growing emphasis on cross-generator generalization and robustness to post-processing. Early and representative pipelines treat detection as supervised real/fake classification using convolutional detectors and forensic cues \cite{wang2020cnn}, while recent benchmarks and large-scale evaluations highlight that distribution shift across generators remains a primary failure mode \cite{zhu2023genimage,ojha2023towards,yan2025sanity}. In parallel, a line of work improves generalization through scaling data and training recipes, leveraging stronger pretrained representations, or introducing alternative training objectives and residual/representation signals \cite{cozzolino2024raising,chen2024drct,chu2025fire,luo2024lare}. Despite steady progress, most methods are still optimized for aggregate accuracy, and provide limited insight into where decisive evidence resides or which evidence is transferable beyond the training generators.

\subsection{Diffusion Representations as Generation-Trace References and Alignment Metrics}
Diffusion models expose rich internal representations that are useful for correspondence and perception 
\cite{tang2023emergent,stracke2025cleandift}, making them a promising 
\emph{generation-trace reference space} for forensic analysis. 
Recent studies further suggest non-trivial correspondence between diffusion representations and external encoder spaces 
\cite{zheng2026diffusion,zhang2025both,yu2025representation}. 
Meanwhile, representation-level interpretability studies show that more classifiable pretrained representations can yield more interpretable evidence structures 
\cite{shen2025enhancing}. 
Together, these findings motivate using diffusion features as a structured reference against which discriminative evidence geometry can be compared. 
Concretely, we build diffusion-feature codebooks as a reference: if a detector's evidence geometry is closer to diffusion traces, its induced forensic concepts tend to be more transferable. 
To quantify such geometric agreement, we adopt neighborhood-structure consistency measures (e.g., CKNNA) motivated by the Platonic representation hypothesis 
\cite{huh2024position}, complementing classic representation similarity tools such as CKA/SVCCA 
\cite{kornblith2019similarity,raghu2017svcca}. 
Additional discussion and related work are deferred to the appendix.
\section{Methodology}
\label{sec:method}

\begin{figure*}[!t]
\begin{center}
\centerline{\includegraphics[width=\columnwidth*2]{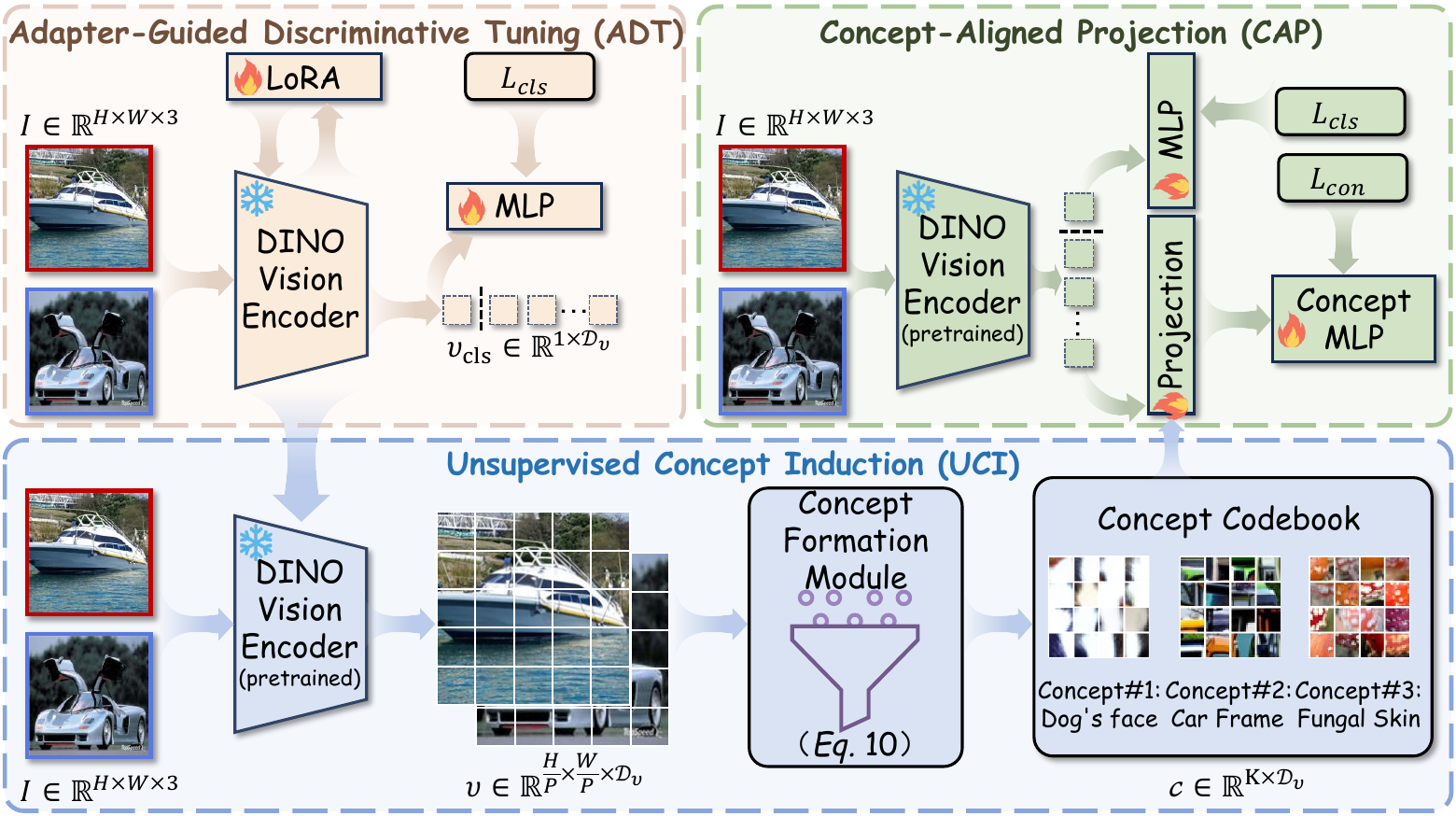}}
\caption{\textbf{Overview of forensic concept learning (\cref{sec:3.1}).} We first perform adapter-guided discriminative tuning (ADT) by inserting LoRA into a pretrained DINO encoder and training a CLS-based detector. Using Transformer attribution, we localize patch-level evidence and cluster the corresponding tokens to induce a forensic concept codebook (UCI). Finally, concept-aligned projection (CAP) maps the CLS representation into the learned concept space.}
\label{fig2}
\end{center}
\vskip -0.2in
\end{figure*}

\begin{figure}[!t]
\begin{center}
\centerline{\includegraphics[width=\columnwidth]{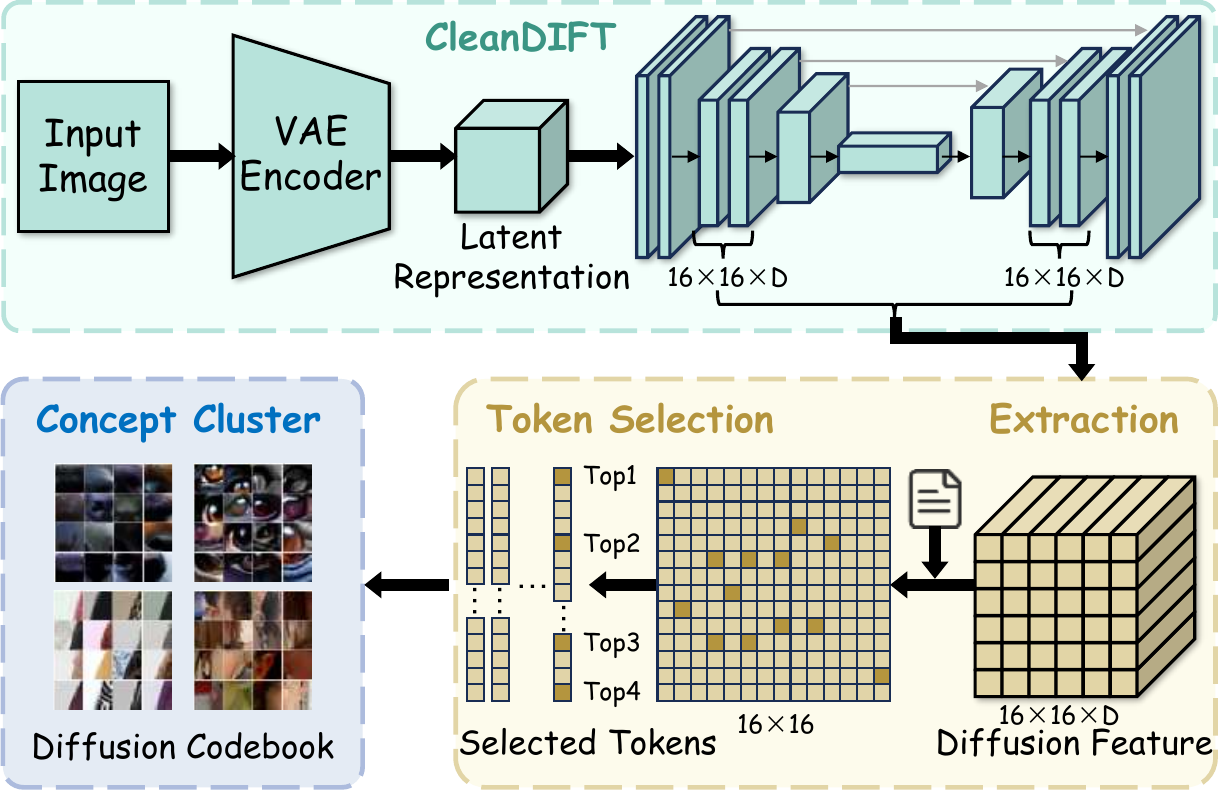}}
\caption{\textbf{CleanDIFT generation-trace reference (\cref{sec:3.2}).}
We extract a $16\times16$ diffusion-token grid $\mathbf{D}^{(l)}(x)$ from a CleanDIFT U-Net layer. Evidence coordinates $\mathcal{I}_b(x)$ select position-aligned tokens for CKNNA computation and diffusion codebook clustering.}
\label{fig3}
\end{center}
\vskip -0.2in
\end{figure}

\begin{figure}[!t]
\begin{center}
\centerline{\includegraphics[width=\columnwidth]{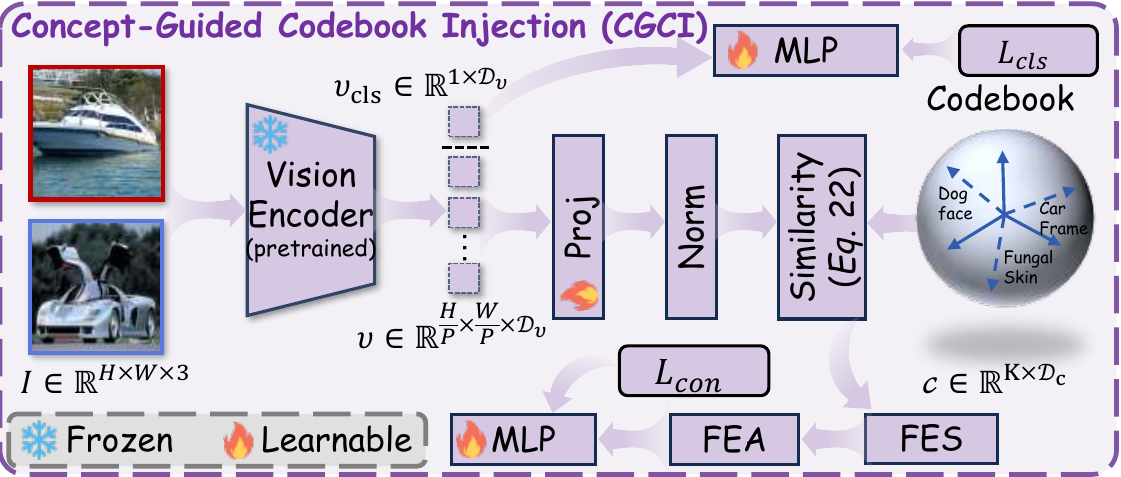}}
\caption{\textbf{Concept-Guided Codebook Injection (CGCI).} CGCI computes normalized patch--concept similarity to a generation-trace codebook, then performs evidence selection (FES) and aggregation (FEA) to form a concept-based prediction alongside the standard CLS pathway.}
\label{fig4}
\end{center}
\vskip -0.2in
\end{figure}

We begin with an empirical observation: a DINO-based detector yields an evidence-aligned concept codebook with visually coherent prototypes that improve detection when used as an explicit concept space, as shown in \cref{fig2}. However, visual coherence alone does not explain whether these concepts reflect \emph{generator traces} or backbone-specific shortcuts. 
To address this, motivated by recent evidence of concept-level correspondence between DINO and diffusion representations, we adopt CleanDIFT features as an external \emph{generation-trace reference} (\cref{fig3}) and quantify evidence--trace agreement via neighborhood-structure consistency (CKNNA). We then validate this analysis through intervention: we construct diffusion codebooks guided by evidence coordinates and inject them into CLIP \cite{radford2021learning} via CGCI (\cref{fig4}), testing whether cross-generator gains correlate with the measured alignment.

\subsection{Decomposing Discriminative Evidence into Forensic Concepts}
\label{sec:3.1}

\paragraph{Hypothesis.}
We hypothesize that a ViT-based detector's decision is driven by a small set of salient patch-level cues that exhibit structured geometry in representation space. These cues can be summarized into a compact forensic concept space. This section describes the construction of this concept space; its transferability is examined in later sections via an independent reference and intervention studies.

\begin{algorithm}[tb]
  \caption{Forensic Concept Learning}
  \label{alg:adt_uci_cap}
  \begin{algorithmic}
    \STATE {\bfseries Input:} Dataset $\mathcal{D}$; pretrained encoder $f_{\theta}$; evidence budget $k$; number of concepts $K$; weight $\lambda$
    \STATE {\bfseries Output:} Codebook $\mathbf{C}$; evidence coordinates $\{\mathcal{I}(x)\}$; projection $\mathbf{W}_c$; heads $g$ and $h$

    \STATE \textbf{Stage 1: ADT.}
    \STATE Freeze $\theta$; insert LoRA adapters; train by minimizing $\mathcal{L}_{\mathrm{cls}}$
    \STATE Freeze the tuned detector for UCI

    \STATE \textbf{Stage 2: UCI.}
    \STATE Initialize $\mathcal{U}\leftarrow \emptyset$
    \FOR{$(x,y)\in\mathcal{D}$}
      \STATE Extract patch tokens $\mathbf{V}(x)$
      \STATE Compute attribution $\mathbf{a}(x)\in\mathbb{R}^{M}$ using Eq.~\eqref{eq:relevance_head}--\eqref{eq:rollout}
      \STATE $\mathcal{I}(x)=\mathrm{TopK}(\mathbf{a}(x),k)$
      \STATE $\mathcal{U}\leftarrow \mathcal{U}\cup\{\mathbf{V}_j(x)\mid j\in\mathcal{I}(x)\}$
    \ENDFOR
    \STATE $\mathbf{C} \leftarrow K\text{-means}(\mathcal{U})$

    \STATE \textbf{Stage 3: CAP.}
    \STATE Initialize $\mathbf{W}_c \leftarrow \mathbf{C}^{\top}$; attach concept head $h$
    \STATE Freeze the backbone; optimize $(g,\mathbf{W}_c,h)$ with $\mathcal{L}=\mathcal{L}_{\mathrm{cls}}+\lambda\mathcal{L}_{\mathrm{con}}$
  \end{algorithmic}
\end{algorithm}

\paragraph{Adapter-Guided Discriminative Tuning (ADT).}
We adapt a pretrained DINOv3 ViT encoder \cite{simeoni2025dinov3} into a discriminative detector while preserving representation geometry for evidence analysis (\cref{fig2}, top-left).
Given a training set $\mathcal{D}=\{(x_i,y_i)\}_{i=1}^{N}$ with $y_i\in\{0,1\}$, we start from a pretrained encoder $f_{\theta}$. For an image $x$, the encoder outputs a CLS token and patch tokens:
\begin{equation}
f_{\theta}(x)=\bigl(\mathbf{z}_{\mathrm{cls}}(x),\, \mathbf{V}(x)\bigr),\quad \mathbf{V}(x)\in\mathbb{R}^{M\times d}.
\label{eq:tokens}
\end{equation}
To reduce representation drift, we freeze the backbone parameters $\theta$, insert LoRA~\cite{hu2022lora} adapters into all Transformer blocks, and train only the LoRA parameters and a lightweight classifier head $g(\cdot)$:
\begin{align}
\hat{y}_{\mathrm{cls}}(x)&=g\bigl(\mathbf{z}_{\mathrm{cls}}(x)\bigr), \label{eq:cls_pred}\\
\mathcal{L}_{\mathrm{cls}}&=-y\log\sigma(\hat{y}_{\mathrm{cls}})-(1{-}y)\log(1{-}\sigma(\hat{y}_{\mathrm{cls}})),
\label{eq:cls_loss}
\end{align}
where $\sigma$ denotes the sigmoid function. This yields a discriminative detector while keeping patch-token representations stable for subsequent evidence analysis.

\paragraph{Unsupervised Concept Induction (UCI).}
Given the ADT-tuned detector, we induce a concept codebook from localized evidence (\cref{fig2}, bottom).
We identify decision-critical patches via gradient-based Transformer attribution. To avoid gradient saturation, we use the logit objective:
\begin{equation}
\hat{y}_{t}(x)=(2y-1)\,\hat{y}_{\mathrm{cls}}(x),
\label{eq:targetlogit}
\end{equation}
ensuring consistent attribution direction for both real ($y=0$) and fake ($y=1$) samples.

Let $\mathbf{A}^{(l,h)}\in\mathbb{R}^{(M+1)\times(M+1)}$ be the attention matrix of head $h$ in layer $l$. We define gradient-weighted relevance:
\begin{equation}
\mathbf{R}^{(l,h)}=\mathrm{ReLU}\!\left(\frac{\partial \hat{y}_{t}(x)}{\partial \mathbf{A}^{(l,h)}}\odot \mathbf{A}^{(l,h)}\right).
\label{eq:relevance_head}
\end{equation}
Head-averaging with residual connections yields:
\begin{equation}
\tilde{\mathbf{R}}^{(l)}=\mathrm{Norm}\!\left(\frac{1}{H}\sum_{h=1}^{H}\mathbf{R}^{(l,h)}+\mathbf{I}\right),
\label{eq:relevance_layer}
\end{equation}
where $\mathrm{Norm}(\cdot)$ performs row-wise normalization and $\mathbf{I}$ is the identity matrix. The attention-rollout relevance is:
\begin{equation}
\mathbf{R}(x)=\prod_{l=1}^{L}\tilde{\mathbf{R}}^{(l)}.
\label{eq:rollout}
\end{equation}
We define the attribution score of patch $j$ as the CLS-to-$j$ relevance $a_j(x)=\mathbf{R}_{0,j}(x)$ for $j=1,\ldots,M$, and select the top-$k$ patches as evidence locations:
\begin{equation}
\mathcal{I}(x)=\mathrm{TopK}\bigl(\mathbf{a}(x),\,k\bigr).
\label{eq:topk}
\end{equation}

We collect evidence-bearing patch tokens across the dataset:
\begin{equation}
\mathcal{U}=\bigcup_{(x,y)\in \mathcal{D}}\left\{\mathbf{V}_j(x)\mid j\in \mathcal{I}(x)\right\}.
\label{eq:U}
\end{equation}
Since these tokens lie in a consistent representation space, we apply $K$-means~\cite{1988Algorithms} clustering to $\mathcal{U}$ with Euclidean distance to obtain a forensic concept codebook:
\begin{equation}
\mathbf{C}=\{\mathbf{c}_1,\ldots,\mathbf{c}_K\}\in\mathbb{R}^{K\times d}.
\label{eq:codebook}
\end{equation}
Each prototype $\mathbf{c}_k$ corresponds to a recurring decision-critical evidence pattern.

\paragraph{Concept-Aligned Projection (CAP).}
We attach a concept branch to the frozen ADT detector (\cref{fig2}, top-right). We initialize a learnable projection $\mathbf{W}_c\in\mathbb{R}^{d\times K}$ from the codebook and map the CLS token into concept space:
\begin{align}
\mathbf{s}(x)&=\mathbf{z}_{\mathrm{cls}}(x)^{\top}\mathbf{W}_c\in\mathbb{R}^{K}, \label{eq:proj}\\
\mathbf{W}_c&\leftarrow \mathbf{C}^{\top}\quad(\text{initialization}).
\label{eq:proj_init}
\end{align}
We freeze the backbone, including the LoRA adapters, and optimize the main classifier head $g$, the concept projection $W_c$, and the concept head $h$:
\begin{align}
\hat{y}_{\mathrm{con}}(x)&=h\bigl(\mathbf{s}(x)\bigr), \label{eq:con_pred}\\
\mathcal{L}_{\mathrm{con}}&=-y\log\sigma(\hat{y}_{\mathrm{con}})-(1{-}y)\log(1{-}\sigma(\hat{y}_{\mathrm{con}})).
\label{eq:con_loss}
\end{align}

The overall training objective is:

\begin{equation}
\mathcal{L}=\mathcal{L}_{\mathrm{cls}}+\lambda\,\mathcal{L}_{\mathrm{con}},
\label{eq:dino_total_loss}
\end{equation}
where $\lambda$ balances the two terms.

\subsection{Generation-Trace Reference from CleanDIFT}
\label{sec:3.2}

\cref{sec:3.1} induces a forensic concept space from discriminative evidence, but its geometry may still capture backbone- or dataset-specific shortcuts. We therefore ground evidence in an external \emph{generation-trace reference}. Prior work suggests structural compatibility between DINO-style features and diffusion representations~\cite{zheng2026diffusion,zhang2025both,yu2025representation}, motivating diffusion traces as a shared reference space. We adopt CleanDIFT~\cite{stracke2025cleandift} as a clean diffusion-trace representation and quantify backbone--trace alignment via neighborhood-structure consistency (CKNNA).

\paragraph{Diffusion trace extraction.}
Given an image $x$, CleanDIFT extracts a dense diffusion-token grid $\mathbf{D}^{(l)}(x)\in\mathbb{R}^{M\times d_l}$ from U-Net layer $l$, where $M$ is the number of spatial locations and $d_l$ is the token dimension.

\paragraph{Position-aligned pairing.}
For a discriminative backbone $b$, we reuse its evidence coordinates $\mathcal{I}_b(x)$ from \cref{sec:3.1}, where each index $j\in\mathcal{I}_b(x)\subset\{1,\dots,M\}$ corresponds to a location on a $16\times16$ grid.
To ensure index-level alignment, we standardize backbone features to $16\times16$ resolution.
For each selected location $j$, we form a paired representation:
\begin{align}
\mathbf{p}_{x,j} &= \mathbf{V}^{b}_{j}(x), \label{eq:backbone_token}\\
\mathbf{q}_{x,j} &= \mathbf{D}^{(l)}_{j}(x), \quad j \in \mathcal{I}_b(x),
\label{eq:representation}
\end{align}
where $\mathbf{p}_{x,j}\in\mathbb{R}^{d_b}$ and $\mathbf{q}_{x,j}\in\mathbb{R}^{d_l}$.
We collect all pairs into $\mathcal{P}=\{u\}$, where $u\equiv(x,j)$ identifies a sampled position, and denote the embeddings as $\mathbf{p}_u$ and $\mathbf{q}_u$.

\paragraph{CKNNA alignment.}
We compute nearest neighbors within each space separately using cosine distance after $\ell_2$-normalization.
Let $k_{\mathrm{NN}}$ denote the neighborhood size.
For each paired sample $u\in\mathcal{P}$, let $\mathcal{N}^{p}(u)$ and $\mathcal{N}^{q}(u)$ denote its $k_{\mathrm{NN}}$-nearest-neighbor sets in the backbone and diffusion spaces, respectively:
\begin{equation}
\mathrm{CKNNA}_{k_{\mathrm{NN}}}(b,l)
=
\frac{1}{|\mathcal{P}|}
\sum_{u \in \mathcal{P}}
\frac{\left|\mathcal{N}^{p}(u)\cap\mathcal{N}^{q}(u)\right|}{k_{\mathrm{NN}}}.
\label{eq:cknna}
\end{equation}
A larger CKNNA indicates stronger neighborhood consistency between backbone evidence and diffusion traces.

\paragraph{Coords-guided diffusion codebook.}
Independently of CKNNA, we cluster sampled diffusion tokens at evidence locations to obtain a diffusion codebook for injection:
\begin{align}
\mathcal{U}^{(l)}_{b}
&=
\bigcup_{x\in\mathcal{D}}
\left\{
\mathbf{D}^{(l)}_{j}(x)\mid j \in \mathcal{I}_b(x)
\right\}, \label{eq:coords_tokens}\\
\mathbf{R}^{(l)}_{b}
&=
\mathrm{KMeans}\!\left(\mathcal{U}^{(l)}_{b};\,K_r\right).
\label{eq:coords_codebook}
\end{align}
This codebook $\mathbf{R}^{(l)}_{b}$ is used for injection in \cref{sec:3.3}.

\subsection{Concept-Guided Codebook Injection (CGCI)}
\label{sec:3.3}

We now validate through intervention: we inject diffusion-derived codebooks into a target backbone and test whether the gains correlate with the measured alignment from \cref{sec:3.2}.

Given an image $x$, a target backbone outputs patch tokens $\mathbf{X}(x)\in\mathbb{R}^{B\times N\times D_x}$ and a global token $\mathbf{z}_{\mathrm{cls}}(x)$.
Let $\mathbf{C}\in\mathbb{R}^{K\times D_c}$ denote a diffusion-derived codebook whose rows represent generation-trace prototypes.
CGCI introduces a concept branch that maps patch tokens to the codebook space, scores patches by concept responses, selects informative patches, and aggregates them for classification.

\paragraph{Codebook-space projection.}
We project patch tokens into the codebook space and compute normalized similarity:
\begin{align}
\mathbf{Q}&=\mathbf{X}\mathbf{W}_q, \quad
\hat{\mathbf{Q}}=\ell_2\text{-norm}(\mathbf{Q}), \quad
\hat{\mathbf{C}}=\ell_2\text{-norm}(\mathbf{C}), \label{eq:proj_norm}\\
\mathbf{S}&=\frac{1}{\tau}\hat{\mathbf{Q}}\hat{\mathbf{C}}^{\top}\in\mathbb{R}^{B\times N\times K},
\label{eq:similarity}
\end{align}
where $\mathbf{W}_q\in\mathbb{R}^{D_x\times D_c}$ is a learnable projection and $\tau$ is a temperature.

\paragraph{Forensic Evidence Scoring (FES).}
We obtain an evidence score per patch by averaging its top-$r$ concept responses:
\begin{equation}
\mathrm{score}_{n,i}
=\frac{1}{r}\sum_{t=1}^{r} S_{n,i,(t)},
\label{eq:score}
\end{equation}
where $S_{n,i,(t)}$ is the $t$-th largest entry in $\mathbf{S}_{n,i,:}$.
This favors patches with concentrated responses to a small subset of prototypes.
We select the top-$m$ patches:
\begin{equation}
\mathcal{J}_n=\mathrm{TopM}(\mathrm{score}_{n,:},\,m).
\label{eq:select_topm}
\end{equation}

\paragraph{Forensic Evidence Aggregator (FEA).}
FEA aggregates selected patches into a global evidence vector using softmax-normalized weights:
\begin{align}
\mathbf{w}^{(n)}&=\mathrm{softmax}\!\left(\mathrm{score}^{(n)}_{\mathrm{sel}}/\tau_w\right), \label{eq:fea_weight}\\
\mathbf{g}^{(n)}&=\sum_{i=1}^{m} w^{(n)}_{i}\,\mathbf{X}^{(n)}_{\mathrm{sel},i},
\label{eq:fea}
\end{align}
where $\mathbf{X}^{(n)}_{\mathrm{sel}}$ contains the selected patch tokens and $\mathbf{g}^{(n)}\in\mathbb{R}^{D_x}$ is the aggregated evidence vector.

\paragraph{Training objective.}
The concept branch predicts $\hat{\mathbf{y}}_{\mathrm{con}}(x)=\mathrm{MLP}_{\mathrm{con}}(\mathbf{g})$ and the main branch predicts $\hat{\mathbf{y}}_{\mathrm{cls}}(x)=\mathrm{MLP}_{\mathrm{cls}}(\mathbf{z}_{\mathrm{cls}})$.
We optimize a joint objective:
\begin{equation}
\mathcal{L}
=\mathcal{L}_{\mathrm{BCE}}(\hat{\mathbf{y}}_{\mathrm{cls}},\mathbf{y})
+\lambda\,\mathcal{L}_{\mathrm{BCE}}(\hat{\mathbf{y}}_{\mathrm{con}},\mathbf{y}),
\label{eq:injection_loss}
\end{equation}
where $\mathcal{L}_{\mathrm{BCE}}$ is the binary cross-entropy loss and $\lambda$ balances the two terms.

Across backbones $b$, we compare CGCI gains against $\mathrm{CKNNA}_{k_{\mathrm{NN}}}(b,l)$ measured in \cref{sec:3.2}. A consistent correlation provides evidence that backbone--trace alignment predicts concept transfer effectiveness.

\begin{table*}[t]
\centering
\setlength{\tabcolsep}{4pt}
\renewcommand{\arraystretch}{1.08}
\caption{\textbf{Benchmarking cross-method generalization on GenImage in terms of Accuracy (\%).}
We follow \cite{zhu2023genimage} and use SDv1.4 as the training set and the others as the test sets. 
Baseline results are directly cited from \cite{yan2025orthogonal}.
We report Accuracy (\%) following \cite{chen2024drct}.
}
\label{tab:genimage_crossmethod_acc}
\resizebox{\textwidth}{!}{%
\begin{tabular}{l l c c c c c c c c c}
\toprule
Methods & Venues & Midjourney & SDv1.4 & SDv1.5 & ADM & GLIDE & Wukong & VQDM & BigGAN & Mean \\
\midrule
ResNet-50 \cite{he2016deep}      & CVPR 2016 & 54.9 & 99.9 & 99.7 & 53.5 & 61.9 & 98.2 & 56.6 & 52.0 & 72.1 \\
DeiT-S \cite{touvron2021training}    & ICML 2021 & 55.6 & 99.9 & 99.8 & 49.8 & 58.1 & 98.9 & 56.9 & 53.5 & 71.6 \\
Swin-T \cite{liu2021swin}       & ICCV 2021 & 62.1 & 99.9 & 99.8 & 49.8 & 67.6 & 99.1 & 62.3 & 57.6 & 74.8 \\
CNNSpot \cite{wang2020cnn}    & CVPR 2020 & 52.8 & 96.3 & 95.9 & 50.1 & 39.8 & 78.6 & 53.4 & 46.8 & 64.2 \\
Spec \cite{zhang2019detecting}        & WIFS 2019 & 52.0 & 99.4 & 99.2 & 49.7 & 49.8 & 94.8 & 55.6 & 49.8 & 68.8 \\
F3Net \cite{qian2020thinking}        & ECCV 2020 & 50.1 & 99.9 & \textbf{99.9} & 49.9 & 50.0 & \textbf{99.9} & 49.9 & 49.9 & 68.7 \\
GramNet \cite{liu2020global}      & CVPR 2020 & 54.2 & 99.2 & 99.1 & 50.3 & 54.6 & 98.9 & 50.8 & 51.7 & 69.9 \\
UnivFD \cite{ojha2023towards}       & CVPR 2023 & 91.5 & 96.4 & 96.1 & 58.1 & 73.4 & 94.5 & 67.8 & 57.7 & 79.5 \\
NPR \cite{tan2024rethinking}          & CVPR 2024 & 81.0 & 98.2 & 97.9 & 76.9 & 89.8 & 96.9 & 84.1 & 84.2 & 88.6 \\
FreqNet \cite{tan2024frequency}      & AAAI 2024 & 89.6 & 98.8 & 98.6 & 66.8 & 86.5 & 97.3 & 75.8 & 81.4 & 86.8 \\
FatFormer \cite{liu2024forgery}     & CVPR 2024 & 92.7 & \textbf{100.0} & \textbf{99.9} & 75.9 & 88.0 & \textbf{99.9} & \textbf{98.8} & 55.8 & 88.9 \\
DRCT \cite{chen2024drct}        & ICML 2024 & 91.5 & 95.0 & 94.4 & \textbf{79.4} & 89.2 & 94.7 & 90.0 & 81.7 & 89.5 \\
AIDE \cite{yan2025sanity}        & ICLR 2025 & 79.4 & 99.7 & 99.8 & 78.5 & 91.8 & 98.7 & 80.3 & 66.9 & 86.9 \\
Effort \cite{yan2025orthogonal} & ICML 2025 & 82.4 & 99.8 & 99.8 & 78.7 & \textbf{93.3} & 97.4 & 91.7 & 77.6 & 91.1 \\
\midrule
\textbf{Ours}                    & --        & \textbf{95.0} & 99.6 & 99.4 & 69.2 & 85.1 & 99.6 & 94.3 & \textbf{94.1} & \textbf{92.0} \\
\bottomrule
\end{tabular}%
}
\end{table*}

\begin{table}[t]
\centering
\small
\setlength{\tabcolsep}{8pt}
\renewcommand{\arraystretch}{1.08}
\caption{\textbf{GenImage cross-generator Accuracy (\%) with and without diffusion-codebook injection.}
Models are trained on the SDv1.4 training split and evaluated on the official GenImage test splits. CLIP (w/o inj.) is the baseline detector, while CLIP (w/ codebook inj.) injects the CleanDIFT-derived diffusion codebook. Parentheses report the absolute $\Delta$Acc relative to the baseline.}
\label{tab:clip_vs_concept_acc}
\begin{tabular}{lcc}
\toprule
Generator & CLIP (w/o inj.) & CLIP (w/ codebook inj.) \\
\midrule
Midjourney & 70.4 & 85.9 (+15.5) \\
SDv1.4     & 99.9 & 99.0 (-0.9) \\
SDv1.5     & 99.8 & 98.8 (-1.1) \\
ADM        & 58.1 & 63.3 (+5.1) \\
GLIDE      & 91.7 & 95.3 (+3.6) \\
Wukong     & 99.0 & 98.4 (-0.6) \\
VQDM       & 76.9 & 84.4 (+7.5) \\
BigGAN     & 74.1 & 80.6 (+6.5) \\
\midrule
Mean       & 83.7 & 88.2 (+4.5) \\
\bottomrule
\end{tabular}
\end{table}

\begin{table*}[!t]
\centering
\caption{\textbf{Backbone--diffusion alignment and codebook injection.}
\textbf{Left:} $100\times$ CKNNA between backbone evidence and CleanDIFT traces from layers s3--s5 and us6--us8 ($k_{\mathrm{NN}}{=}10$).
\textbf{Right:} GenImage cross-generator accuracy (\%) of CLIP with codebooks from different sources.
\textbf{Bold} indicates the best result per column.
\emph{Abbrev.:} MJ = Midjourney; SD4/SD5 = Stable Diffusion v1.4/v1.5.}
\label{tab:cknna_and_genimage}

\newcommand{\LeftStretch}{1.32}

\begin{minipage}[t]{0.38\textwidth}
\vspace{0pt}
\centering
\small
\setlength{\tabcolsep}{3pt}
\renewcommand{\arraystretch}{\LeftStretch}
\begin{tabular}{lcccccc}
\toprule
\multirow{2}{*}{Backbone} & \multicolumn{3}{c}{Down} & \multicolumn{3}{c}{Up} \\
\cmidrule(lr){2-4}\cmidrule(lr){5-7}
& s3 & s4 & s5 & us6 & us7 & us8 \\
\midrule
DINOv3          & 5.5  & \textbf{8.8}  & \textbf{9.5}  & \textbf{24.4} & \textbf{17.2} & \textbf{18.9} \\
CLIP          & 4.0  & 5.6  & 5.6  & 8.9  & 7.2  & 8.9  \\
DeiT          & 4.5  & 6.5  & 6.6  & 14.7 & 10.1 & 10.4 \\
Swin-T        & 2.5  & 3.5  & 3.6  & 7.0  & 5.5  & 6.1  \\
ResNet        & 0.4  & 4.5  & 4.2  & 3.2  & 3.3  & 3.5  \\
EfficientNet  & \textbf{5.9} & 7.2  & 6.7  & 5.6  & 5.8  & 7.1  \\
\bottomrule
\end{tabular}
\end{minipage}
\hfill
\begin{minipage}[t]{0.60\textwidth}
\vspace{0pt}
\centering
\small
\setlength{\tabcolsep}{1.8pt}
\renewcommand{\arraystretch}{1.05}
\begin{tabular*}{\linewidth}{@{\extracolsep{\fill}}lccccccccc}
\toprule
Source & MJ & SD4 & SD5 & ADM & GLIDE & Wukong & VQDM & BigGAN & Mean \\
\midrule
CLIP w/o inj &
70.4 & \textbf{99.9} & \textbf{99.8} & 58.1 & 91.7 & \textbf{99.0} & 76.9 & 74.1 & 83.7 \\
\midrule
\multicolumn{10}{l}{\emph{Injecting codebooks into CLIP}} \\
DINOv3 &
74.2 & 99.8 & 99.6 & 60.1 & 92.2 & 98.6 & 80.6 & 78.4 & 85.4 \\
CLIP &
72.1 & 99.8 & \textbf{99.8} & 60.5 & 91.0 & 98.5 & 79.4 & 75.4 & 84.6 \\
DeiT &
70.2 & 99.8 & \textbf{99.8} & 60.1 & 91.5 & 98.9 & 79.9 & 79.3 & 85.0 \\
Swin-T &
70.2 & 99.8 & \textbf{99.8} & 60.1 & 90.2 & 98.7 & 77.9 & 77.5 & 84.3 \\
ResNet &
73.3 & 99.8 & \textbf{99.8} & 60.6 & 91.3 & 98.5 & 79.6 & 75.6 & 84.8 \\
EfficientNet &
76.1 & 99.4 & 99.3 & \textbf{64.9} & 93.8 & 98.8 & \textbf{84.5} & \textbf{83.5} & 87.5 \\
CleanDIFT &
\textbf{85.9} & 99.0 & 98.8 & 63.3 & \textbf{95.3} & 98.4 & 84.4 & 80.6 & \textbf{88.2} \\
\bottomrule
\end{tabular*}
\end{minipage}
\end{table*}

\begin{figure}[!t]
\begin{center}
\centerline{\includegraphics[width=\columnwidth]{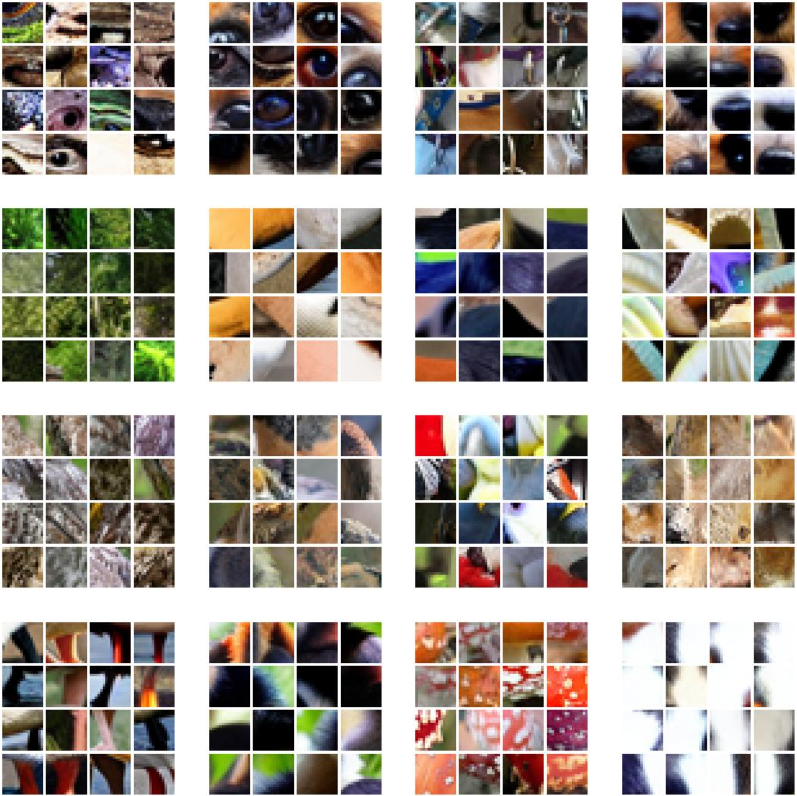}}
\caption{\textbf{DINOv3 concept codebook centers on GenImage-SDv1.4.}
We induce a $K=200$ codebook from attribution-selected evidence patches and visualize concepts by nearest-neighbor patch collages, revealing coherent local cues for real/fake decisions.}
\label{fig:5}
\end{center}
\vskip -0.2in
\end{figure}

\section{Experiment}
\label{sec:experiments}
This section follows a progressive evidence chain.
We first present the main GenImage comparison in \cref{tab:genimage_crossmethod_acc} to show that the proposed DINOv3 concept codebook improves cross-generator detection and yields auditable evidence units. We then justify our \emph{generation-trace reference} by injecting a CleanDIFT-derived codebook into CLIP and observing consistent cross-generator gains in \cref{tab:clip_vs_concept_acc}.
Finally, we test whether CKNNA-based backbone--trace alignment predicts transferability across codebook sources in \cref{tab:cknna_and_genimage}.
We end with qualitative visualizations.

\subsection{Datasets and Evaluation Metrics}
\label{sec:datasets-metrics}

\paragraph{Datasets.}
We conduct experiments on three benchmarks. Our main setting is \textbf{GenImage}~\citep{zhu2023genimage}, a standard cross-generator benchmark with real images from ImageNet~\citep{deng2009imagenet} and fakes from multiple generators (e.g., Midjourney, SD variants, ADM, GLIDE, Wukong, VQDM, and BigGAN).
Following the official protocol, we train on the SD~v1.4 subset and evaluate on the official test subsets; the training set contains 324K images (162K real and 162K SD~v1.4 fakes), and all inputs are resized to $224\times224$.
We further test cross-family transfer on the \textbf{AIGCDetect} GAN-family benchmark~\citep{zhong2023patchcraft} (seven GAN generators) and report results on \textbf{Chameleon}~\citep{yan2025sanity} with stronger distribution shifts.
A detailed cross-method GenImage comparison is reported in \cref{tab:genimage_crossmethod_acc}, while additional GAN-family and Chameleon results are provided in the appendix.

\paragraph{Evaluation metrics.}
We use two metrics throughout: \textbf{ACC} (real-vs-fake accuracy; mean over the official GenImage test splits) and \textbf{CKNNA} (average $k_{\mathrm{NN}}$-nearest-neighbor overlap between position-aligned backbone evidence and CleanDIFT tokens, as defined in \cref{sec:3.2}).

\begin{figure}[!t]
\begin{center}
\centerline{\includegraphics[width=\columnwidth]{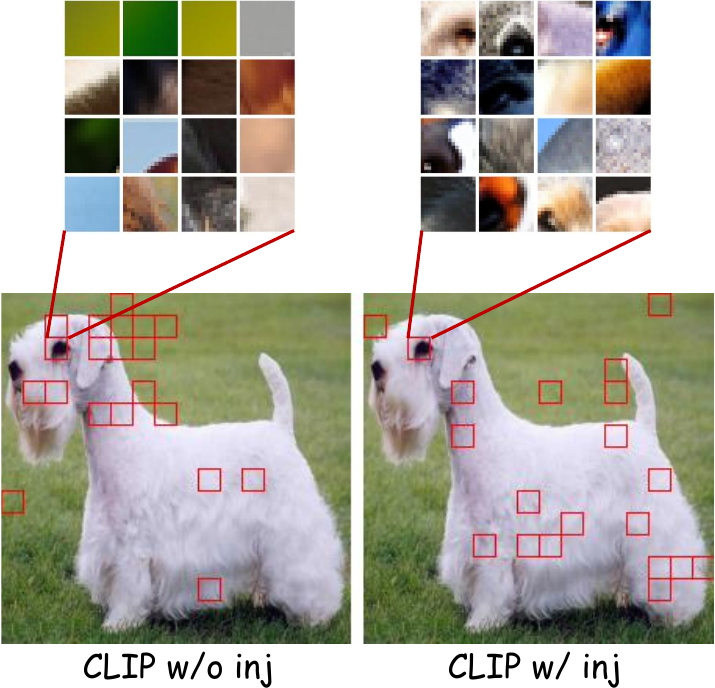}}
\caption{\textbf{Model-attended patches before and after codebook injection.}
Red boxes show the top-$20$ patches on the input image.
\textbf{Left:} CLIP \emph{w/o inj}, ranked by Transformer Attribution.
\textbf{Right:} CLIP \emph{w/ inj}, ranked by codebook-similarity (concept-response) scores.
Top panels visualize each selected patch’s nearest codebook cluster center by displaying the $16$ closest patches to that prototype.}
\label{fig8}
\end{center}
\vskip -0.2in
\end{figure}

\subsection{Models and Implementation Details}
\paragraph{Implementation details.}
Our DINOv3 detector follows the three-stage pipeline in \cref{sec:3.1}: ADT applies LoRA tuning with Adam ($\mathrm{lr}=1{\times}10^{-4}$); UCI freezes Stage-I, selects top-$k$ evidence patches on the $16{\times}16$ grid ($k{=}20$) via Transformer attribution, and clusters evidence tokens with KMeans ($K{=}200$) to form the concept codebook; CAP trains the concept branch (no LoRA) with Adam ($\mathrm{lr}=2{\times}10^{-5}$) and $\lambda{=}1$, while the plain DINOv3 baseline uses the same CAP setting for ablation.
For generation-trace analysis (\cref{sec:3.2}), we extract pretrained CleanDIFT features with $256{\times}256$ inputs to match the $16{\times}16$ grid, then sample diffusion tokens at backbone evidence coordinates for CKNNA and to build diffusion codebooks for injection.
All experiments are implemented in PyTorch and run on a single NVIDIA A100 GPU (40GB).

\subsection{Generation-Codebook Injection Improves CLIP}
\label{sec:exp_injection_effect}
We first validate that diffusion-derived codebook injection yields consistent cross-generator improvements on CLIP.
\cref{tab:clip_vs_concept_acc} reports GenImage performance with and without injection: CleanDIFT injection increases the mean ACC from 83.7 to 88.2, with gains on most test generators.
This establishes injection as an effective intervention, independent of any alignment analysis.

\subsection{Alignment Predicts Which Codebooks Transfer}
\label{sec:exp_injection_tracks_alignment}
We study when codebook injection helps and which sources transfer best to CLIP.
We measure evidence--trace agreement using CKNNA between position-aligned backbone evidence tokens and CleanDIFT tokens on the $16{\times}16$ grid (\cref{sec:3.2}).
The left half of \cref{tab:cknna_and_genimage} shows that DINO is most aligned with diffusion traces, while other backbones are less aligned, supporting our introductory hypothesis that discriminative ViT representations exhibit non-trivial structural compatibility with diffusion features.
The right half shows that injecting more aligned codebooks generally yields larger cross-generator gains, suggesting CKNNA predicts transferability.
The CleanDIFT-derived codebook provides the strongest improvement, consistent with \cref{tab:clip_vs_concept_acc}.
EfficientNet outperforms DINO despite lower CKNNA; within CNNs, gains still follow the CKNNA ordering.

\begin{figure}[!t]
\begin{center}
\centerline{\includegraphics[width=\columnwidth]{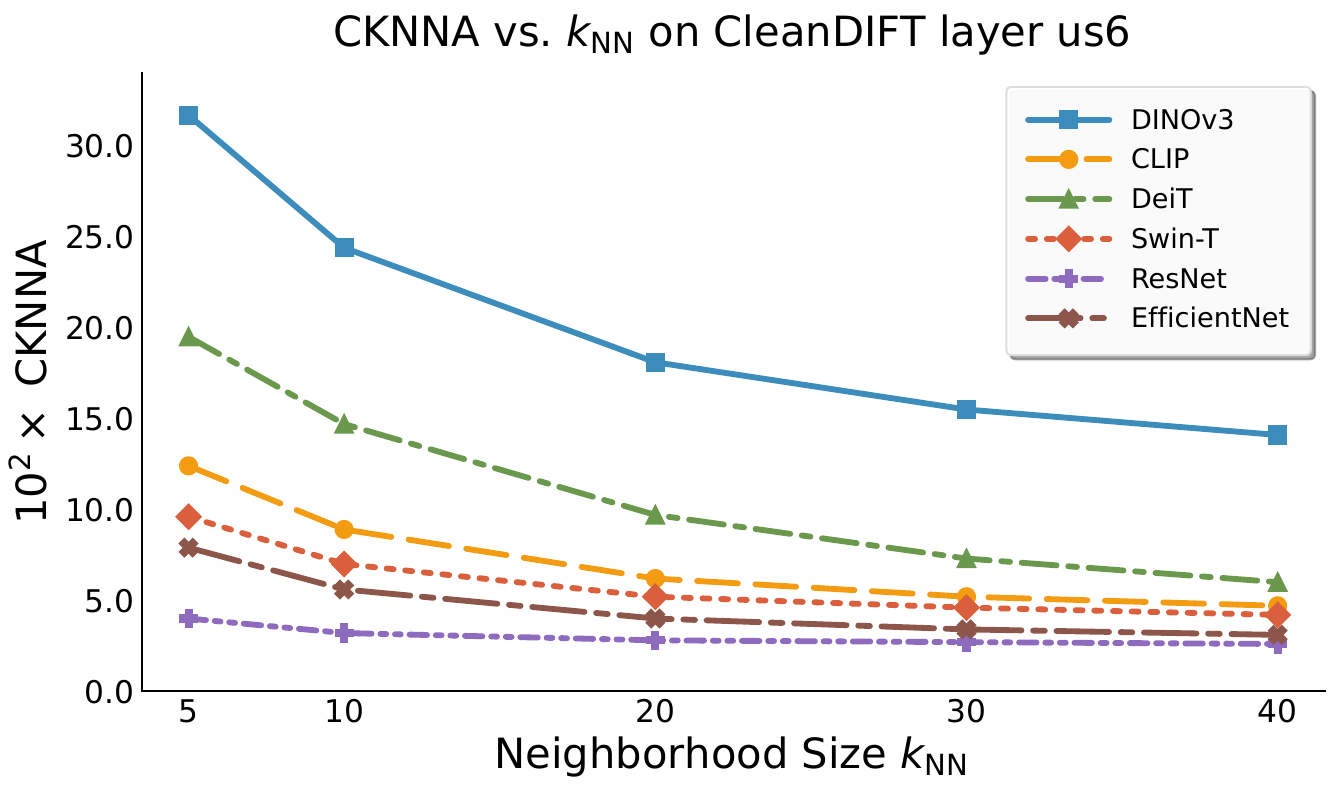}}
\caption{\textbf{CKNNA sensitivity to neighborhood size on CleanDIFT us6.}
For each discriminative backbone, we compute CKNNA between attribution-selected backbone evidence tokens and position-aligned CleanDIFT us6 tokens, varying the neighborhood size $k_{\mathrm{NN}}\in\{5,10,20,30,40\}$.
Values are scaled by $100$ for readability.}
\label{fig:cknna_vs_k}
\end{center}
\vskip -0.2in
\vspace{-1em}
\end{figure}

\paragraph{Sensitivity of CKNNA to neighborhood size.}

\cref{fig:cknna_vs_k} evaluates the sensitivity of CKNNA to the neighborhood size $k_{\mathrm{NN}}$ on the CleanDIFT us6 layer.
CKNNA decreases as $k_{\mathrm{NN}}$ increases from $5$ to $40$, since larger neighborhoods include less similar samples and are less consistently ordered across representation spaces.
Despite this decrease, the relative backbone ranking remains largely stable: DINOv3 shows the strongest alignment with diffusion traces, followed by DeiT and CLIP, while Swin-T, ResNet, and EfficientNet show lower agreement.
Thus, our alignment conclusion is not tied to a specific neighborhood size, and we use $k_{\mathrm{NN}}{=}10$ by default.

\begin{table*}[t]
\centering
\small
\setlength{\tabcolsep}{4pt}
\renewcommand{\arraystretch}{1.08}
\caption{\textbf{Post-processing robustness results.}
We report accuracy (\%) under common image perturbations, including JPEG compression, Gaussian blur, Gaussian noise, color jitter, and resizing. 
\textbf{Bold} indicates the best result in each column.}
\label{tab:post_processing_robustness}
\resizebox{\textwidth}{!}{%
\begin{tabular}{lcccccccccc}
\toprule
\multirow{2}{*}{Method} 
& \multirow{2}{*}{Original}
& \multicolumn{2}{c}{JPEG Compression}
& \multicolumn{2}{c}{Gaussian Blur}
& \multicolumn{2}{c}{Gaussian Noise}
& \multicolumn{2}{c}{Color Jitter}
& \multirow{2}{*}{Resize} \\
\cmidrule(lr){3-4}
\cmidrule(lr){5-6}
\cmidrule(lr){7-8}
\cmidrule(lr){9-10}
& 
& QF=75 & QF=70 
& $\sigma=1.0$ & $\sigma=2.0$
& $\sigma=0.1$ & $\sigma=0.2$
& contrast=0.2 & brightness=0.2
& $\times0.5$ \\
\midrule
CLIP   
& 83.70 & 50.50 & 50.61 & 83.62 & 77.36 & 59.84 & 51.23 & 83.22 & 80.35 & 80.10 \\
Effort 
& 91.10 & 63.37 & 61.75 & 78.50 & 73.39 & 65.87 & 60.89 & 89.73 & 89.44 & 81.39 \\
\textbf{Ours}   
& \textbf{92.00} & \textbf{91.83} & \textbf{91.82} 
& \textbf{92.27} & \textbf{90.56} 
& \textbf{87.22} & \textbf{83.02} 
& \textbf{92.48} & \textbf{91.37} & \textbf{91.49} \\
\bottomrule
\end{tabular}%
}
\vspace{-1em}
\end{table*}

\subsection{Ablation Study}
\label{sec:exp_ablation}
We analyze the contribution of \emph{concept-space prediction} in \cref{tab:ablation_concept_mean}. 
All results follow the same evaluation protocol: AIDE and Effort use released checkpoints, while other backbone baselines are reproduced using the same training split. 
Adding CAP to DINOv3 improves the mean accuracy from 87.2 to 88.8 across GenImage, GAN-family, and Chameleon, suggesting that explicit concept-space evidence provides a more transferable decision signal under generator and distribution shifts. Additional ablations are provided in the appendix.

\subsection{Robustness to Post-processing Perturbations}
\label{sec:exp_robustness}
Real-world images are often post-processed, which may weaken forensic traces. 
We evaluate robustness under JPEG compression, Gaussian blur, Gaussian noise, color jitter, and resizing. 
As shown in \cref{tab:post_processing_robustness}, our method consistently outperforms CLIP and Effort across all perturbations. 
For example, it retains 91.83\% accuracy under JPEG compression (QF=75) and 83.02\% under Gaussian noise ($\sigma=0.2$), substantially surpassing both baselines. 
This suggests that our forensic concepts capture stable evidence patterns beyond fragile pixel-level artifacts.

\subsection{Visualization}
\label{sec:exp_viz}
We visualize both what the injected model attends to and what the learned concepts represent. \cref{fig8} shows that codebook injection shifts CLIP’s top evidence patches from relatively noisy, hard-to-interpret attribution hotspots to patches that exhibit clearer, more coherent matches to diffusion prototypes (see caption). \cref{fig:5} further visualizes representative DINOv3 concept centers by retrieving nearest-neighbor evidence patches, revealing coherent recurring local cues induced by UCI. More concept visualizations and qualitative analyses are provided in the appendix.

\begin{table}[t]
\centering
\small
\setlength{\tabcolsep}{4pt}
\renewcommand{\arraystretch}{1.08}
\caption{\textbf{Baselines and concept-branch ablation.}
Accuracy (\%) on GenImage, GAN-family, Chameleon, and their mean.
\textbf{Bold} marks the best result.}
\label{tab:ablation_concept_mean}
\begin{tabular}{lcccc}
\toprule
Methods & GenImage & GAN & Chameleon & Mean \\
\midrule
ResNet & 69.2 & 50.7 & 63.7 & 61.2 \\
EfficientNet & 71.5 & 53.6 & 59.7 & 61.6 \\
Swin-T & 73.7 & 57.2 & 61.1 & 64.0 \\
DeiT & 73.1 & 53.1 & 67.7 & 64.6 \\
CLIP & 83.7 & 79.3 & 58.2 & 73.7 \\
AIDE & 86.9 & 70.7 & 62.6 & 73.4 \\
Effort & 91.1 & 86.6 & 63.2 & 80.3 \\
DINOv3 w/o concept & 90.7 & 87.3 & 83.7 & 87.2 \\
DINOv3 w/ concept  & \textbf{92.0} & \textbf{90.1} & \textbf{84.4} & \textbf{88.8} \\
\bottomrule
\end{tabular}
\vspace{-1em}
\end{table}

\section{Conclusion}
In this paper, we show that a detector’s patch evidence can be organized into an explicit forensic concept space that improves cross-generator AIGI detection while keeping decisions inspectable. We induce a compact concept codebook from attribution-selected DINOv3 evidence and train a lightweight concept branch to expose localized, human-auditable cues. To explain transfer across backbones, we use CleanDIFT generation traces as a generator-grounded reference and quantify backbone–trace compatibility with CKNNA. Experiments indicate that higher alignment yields more transferable codebooks and larger gains when injected into CLIP. Finally, generation-trace alignment offers a practical diagnostic for anticipating which concept codebooks will transfer effectively across backbones.

\section*{Impact Statement}
This work aims to improve the reliability and auditability of AI-generated image (AIGI) detection by introducing a concept-level evidence interface and a generation-trace reference for diagnosing transferability across generators and backbones.
A positive impact is to support real-world content authenticity, moderation, and forensic analysis with decisions that are more interpretable and easier to verify, which can help mitigate misinformation and fraud on online platforms.
A potential negative impact is that clearer evidence visualizations and alignment analyses may inform adaptive evasion strategies (e.g., targeted post-processing to weaken specific forensic cues).
We view this as a standard dual-use risk in forensics: making failure modes explicit can also accelerate the development of more robust detectors and more realistic evaluation protocols.

\section*{Acknowledgements}
This work is supported by the National Key Research and Development Program of China (No. 2025YFE0113500), the National Natural Science Foundation of China (No. U22B2051, No. U25B2066, No. 62302411).

\bibliography{example_paper}
\bibliographystyle{icml2026}

\newpage
\appendix
\onecolumn
\section{Appendix}
\begin{figure}[H]
\centering
\includegraphics[width=\columnwidth]{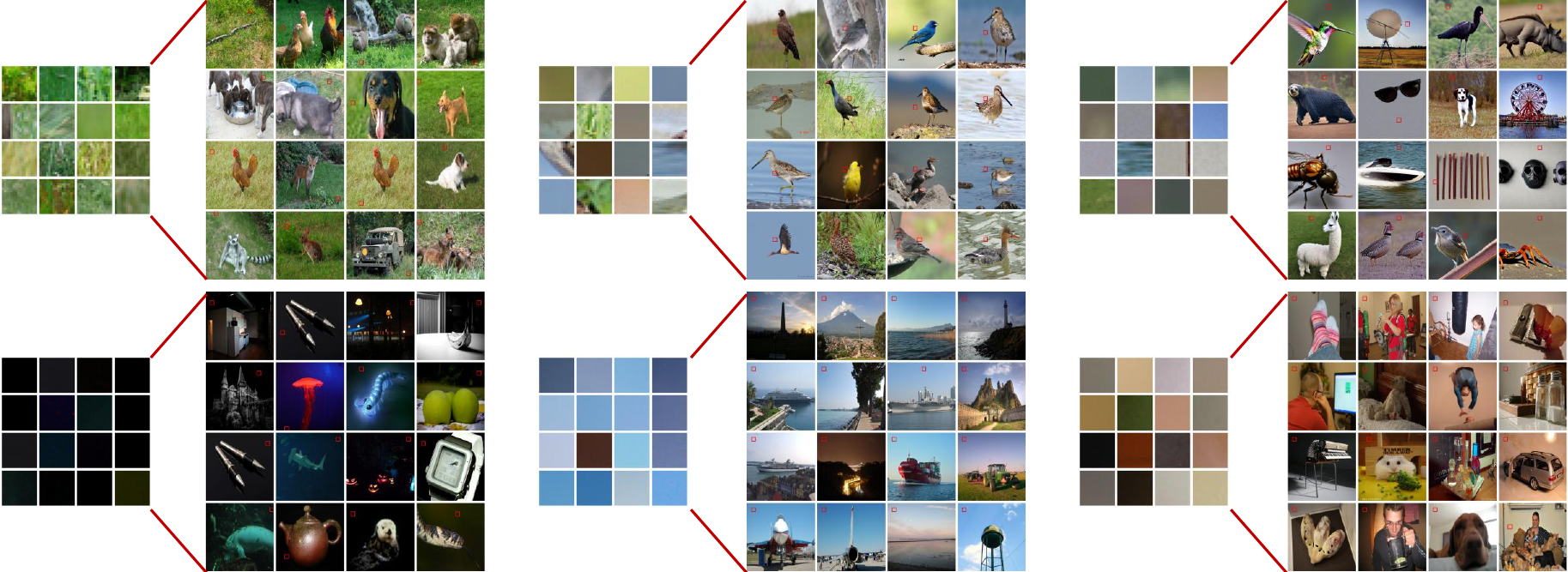}
\caption{\textbf{Stable Diffusion 1.4, Swin-T.}}
\label{fig:codebook_vis_sd14_swin-t}
\end{figure}

This appendix supplements the main paper with additional experimental and implementation details.
We first describe the datasets, evaluation protocols, and implementation settings used in the extended experiments.
We then provide concise explanations of the concept-codebook construction and CGCI procedures in \cref{alg:codebook,alg:cgci}.
Next, we report and discuss additional results on GAN-family and Chameleon benchmarks, mixed-generator training, recent-generator evaluation, post-processing robustness, and hyperparameter ablations.
We also provide extensive codebook-center visualizations in \cref{fig:codebook_vis_sd14_swin-t}--\cref{fig:codebook_vis_stylegan_dino}.
Finally, we discuss limitations and future directions.

\section{Additional Related Work}
\subsection{Explainability and Concept-Based Evidence in Vision Models}
A broad set of explainability methods aim to localize decision evidence, including gradient-based saliency, integrated gradients, and transformer-specific attribution/rollout mechanisms, as well as visualization of patch interactions in ViTs~\cite{sundararajan2017axiomatic,abnar2020quantifying,chefer2021transformer,ma2023visualizing}.
However, localization alone rarely yields reusable forensic units: explanations are often model-specific and do not directly support evidence transfer across heterogeneous backbones.
Complementary studies suggest that task-relevant visual information is often unevenly distributed across activations, layers, and fine-grained regions, as shown in outlier-aware ViT quantization, information-theoretic pruning, and high-resolution reasoning segmentation~\citep{ma2024outlier,zheng2025information,lin2025hrseg}.
This observation is consistent with our view that forensic evidence may reside in localized yet structured representations, motivating us to organize attribution-selected patches into reusable concept units.
Concept-based interpretability instead seeks intermediate, human-auditable units, such as Testing with Concept Activation Vectors (TCAV)~\cite{kim2018interpretability}, concept bottleneck models~\cite{koh2020concept}, and prototype-based reasoning~\cite{chen2019looks}.
Most relevant to our motivation, \cite{shen2025enhancing} studies the relationship between representation classifiability and interpretability, showing that more classifiable pretrained representations can expose more interpretable semantics.
Our work follows this spirit but targets forensic concepts: we convert attribution-localized patches into transferable concept units and use the concept space as an auditable evidence interface for AIGI detection.

\subsection{Transferable Forensic Cues and Evidence Dissection}
Understanding what detectors rely on across generators is crucial for reliable forensics.
Transferable forensic features have been studied as cues that remain predictive under generator shifts~\cite{chandrasegaran2022discovering,ojha2023towards}.
Face forgery detection has also explored cross-domain generalization through learning-to-weight~\citep{sun2021domain}, dual-contrastive learning~\citep{sun2022dual}, and guided Stable-Diffusion-based enhancement~\citep{sun2024diffusionfake}.
Recent works further improve detector reliability from complementary perspectives, including attention-driven evidence modeling~\cite{sun2022information} and uncertainty calibration~\cite{jin2025towards}.
Beyond image-level detection, localized forgery detection shows that forensic cues can be subtle and spatially uneven~\cite{cai2026zooming}, while evasion-oriented studies reveal the vulnerability of current detectors to adaptive generation~\cite{zhou2024stealthdiffusion}.
Together, these studies indicate that reliable detection requires not only strong image-level accuracy but also a clearer understanding of the evidence behind each prediction.
At the same time, many high-performing detectors remain black-box, and existing dissection efforts do not yet provide a general mechanism to (i) organize evidence into reusable concept units, and (ii) transfer such units across heterogeneous backbones with a quantitative notion of alignment.
Our framework addresses this gap by (i) localizing decision-critical patches via transformer attribution, (ii) organizing the resulting evidence tokens into an explicit forensic concept codebook, and (iii) injecting the codebook into heterogeneous backbones for transferable detection.

\section{Datasets and Evaluation Protocols}
This section complements the main paper with dataset-specific notes and the evaluation protocol adopted in our experiments.
We follow the official dataset splits and preprocessing when available, and report binary real/fake classification accuracy (\%) 
unless stated otherwise.

\paragraph{GenImage.}
GenImage is a widely-used cross-generator benchmark for AI-generated image detection, designed to evaluate generalization across different generation engines. 
In our main setting, we train the detector on a designated source generator and test on multiple held-out generators under the official protocol, 
so that the measured performance reflects robustness to generator shift rather than in-distribution fitting.
When the backbone has a fixed input size (e.g., CLIP at $224\times224$), we follow its standard preprocessing; otherwise we use a unified resize/crop setting
consistent with the main paper. \textbf{All CGCI results are reported using the concept head; the CLS head is auxiliary.}

\begin{algorithm}[H]
  \caption{Coords-guided Diffusion Codebook Extraction}
  \label{alg:codebook}
  \begin{algorithmic}
    \STATE {\bfseries Input:} dataset $\mathcal{D}=\{(x_i,y_i)\}_{i=1}^{N}$;
    backbone evidence selector $\mathcal{I}_b(x)\subset\{1,\dots,M\}$, $|\mathcal{I}_b(x)|=k_{\text{ev}}$;
    CleanDIFT tokens $\mathbf{D}^{(l)}(x)\in\mathbb{R}^{M\times d_l}$;
    layer set $\mathcal{L}$; codebook size $K_r$.
    \STATE {\bfseries Output:} codebooks $\{\mathbf{R}^{(l)}_{b}\}_{l\in\mathcal{L}}$.

    \FORALL{$l \in \mathcal{L}$}
      \STATE $\mathcal{U}\leftarrow [\ ]$ \COMMENT{diffusion tokens sampled at evidence coords}
      \FORALL{$(x,y)\in\mathcal{D}$}
        \STATE $\mathcal{I}\leftarrow \mathcal{I}_b(x)$
        \STATE $\mathbf{T}\leftarrow \mathbf{D}^{(l)}(x)$ \COMMENT{$M\times d_l$ diffusion token grid}
        \FORALL{$j\in\mathcal{I}$}
          \STATE append $\mathbf{T}[j]$ to $\mathcal{U}$
        \ENDFOR
      \ENDFOR
      \STATE $\mathbf{R}^{(l)}_{b}\leftarrow \mathrm{KMeans}(\mathcal{U},K_r)$ \COMMENT{$\mathbf{R}^{(l)}_{b}\in\mathbb{R}^{K_r\times d_l}$}
    \ENDFOR

    \STATE {\bfseries return} $\{\mathbf{R}^{(l)}_{b}\}_{l\in\mathcal{L}}$
  \end{algorithmic}
\end{algorithm}

\paragraph{GAN-family benchmark.}
\cref{tab:gan_family_acc} reports cross-generator results on a GAN-family benchmark that covers representative GAN architectures 
(e.g., ProGAN/StyleGAN/StyleGAN2/BigGAN) and image-to-image translation models (e.g., CycleGAN/StarGAN/GauGAN).
Following common practice, we group methods by their training set to avoid confounding training distribution with generalization:
(i) models trained on ProGAN (GAN family) and evaluated on other GAN/translation generators; and 
(ii) models trained on SDv1.4 (diffusion family) and evaluated on the same GAN-family test suite.
This protocol directly probes whether diffusion-derived forensic concepts and generation-trace references can transfer to GAN-generated artifacts.

\begin{algorithm}[t]
  \caption{Concept-Guided Codebook Injection (CGCI)}
  \label{alg:cgci}
  \begin{algorithmic}
    \STATE {\bfseries Input:} image $x$; backbone $f(\cdot)$; diffusion codebook $C$; temperatures $\tau,\tau_w$; hyperparameters $r,m$; loss weight $\lambda$; label $y$
    \STATE {\bfseries Output:} cls-head $\hat{\mathbf{y}}_{\mathrm{cls}}$; concept-head $\hat{\mathbf{y}}_{\mathrm{con}}$; training loss $L$

    \STATE \textit{\textbf{Extract Feature.}}
    \STATE $(z_{\text{cls}}, X) \leftarrow f(x)$ \COMMENT{$X \in \mathbb{R}^{B\times N\times D_{x}}$, $z_{\text{cls}} \in \mathbb{R}^{B\times D_{x}}$}

    \STATE \textit{\textbf{Projection \& Normalization.}}
    \STATE $Q \leftarrow \text{Proj}(X)$
    \STATE $\hat{Q} \leftarrow \text{L2Norm}(Q)$
    \STATE $\hat{C} \leftarrow \text{L2Norm}(C)$

    \STATE \textit{\textbf{Patch-to-Concept Similarity.}}
    \STATE $S \leftarrow \hat{Q}\hat{C}^{\top}/\tau$

    \STATE \textit{\textbf{Forensic Evidence Scoring (FES).}} Applied independently for each sample in the batch.

    \FOR{$n=1$ {\bfseries to} $B$}
      \FOR{$i=1$ {\bfseries to} $N$}
        \STATE $\text{score}_{n,i} \leftarrow \text{Mean}(\text{TopR}(S_{n,i,:}, r))$
      \ENDFOR
    \ENDFOR

    \STATE \textit{\textbf{Top-$m$ Evidence Patch Selection.}} 
    \STATE $\text{idx} \leftarrow \text{TopMIdx}(\text{score}, m)$
    \STATE $X_{\text{sel}} \leftarrow \text{Gather}(X, \text{idx})$
    \STATE $\text{score}_{\text{sel}} \leftarrow \text{Gather}(\text{score}, \text{idx})$

    \STATE \textit{\textbf{Forensic Evidence Aggregator (FEA).}} 
    \STATE $w \leftarrow \text{Softmax}(\text{score}_{\text{sel}}/\tau_w)$
    \STATE $g \leftarrow \sum_{j=1}^{m} w_j \, X_{\text{sel},j}$

    \STATE \textit{\textbf{Dual-Head Prediction.}}
    \STATE $\hat{\mathbf{y}}_{\mathrm{con}} \leftarrow \text{MLP}_{\text{con}}(g)$
    \STATE $\hat{\mathbf{y}}_{\mathrm{cls}} \leftarrow \text{MLP}_{\text{cls}}(z_{\text{cls}})$

    \STATE \textit{\textbf{Joint Training Objective.}}
    \STATE $L \leftarrow \text{BCEWithLogits}(\hat{\mathbf{y}}_{\mathrm{cls}}, y)$
    \STATE \hspace{1.6em}$\;\;\;+\lambda\,\text{BCEWithLogits}(\hat{\mathbf{y}}_{\mathrm{con}}, y)$

    \STATE \textbf{return} $\hat{\mathbf{y}}_{\mathrm{cls}}, \hat{\mathbf{y}}_{\mathrm{con}}, L$
  \end{algorithmic}
\end{algorithm}

\paragraph{Chameleon.}
Chameleon is a challenging, diverse benchmark that aggregates images produced by modern generative pipelines and post-processing mechanisms, 
with substantial content and style variability. Compared to curated single-family benchmarks, Chameleon is intentionally harder and 
better reflects practical deployment conditions where the generative source may be unknown or rapidly evolving.
\cref{tab:chameleon_acc_wide} reports accuracy for a broad set of detectors under the same evaluation setup.

\section{Pseudocode Details}
This section explains the intent and the key design choices behind the algorithms included in the appendix.

\paragraph{Coords-guided Diffusion Codebook Extraction (\cref{alg:codebook}).}
The goal is to construct a generation-trace codebook that is \emph{evidence-aligned} with a given discriminative backbone $b$.
For each image $x$, the backbone provides a set of evidence coordinates $\mathcal{I}_b(x)$ (e.g., top-$k_{\text{ev}}$ patch indices ranked by attribution/evidence).
Given CleanDIFT tokens $\mathbf{D}^{(l)}(x)\in\mathbb{R}^{M\times d_l}$ at a chosen U-Net layer $l$, we sample only the tokens at those evidence coordinates.
Aggregating these sampled diffusion tokens over the dataset yields a compact set $\mathcal{U}$ that focuses on where the detector looks rather than 
uniformly pooling over all patch positions. We then run KMeans with $K_r$ clusters to obtain the layer-wise codebook
$\mathbf{R}^{(l)}_{b}\in\mathbb{R}^{K_r\times d_l}$. This procedure is repeated across the layer set $\mathcal{L}$, producing 
$\{\mathbf{R}^{(l)}_{b}\}_{l\in\mathcal{L}}$.

\paragraph{Concept-Guided Codebook Injection (CGCI, \cref{alg:cgci}).}
CGCI injects a diffusion-derived codebook $C$ into a target backbone $f(\cdot)$ via a lightweight concept branch, without modifying the backbone architecture.
Given patch tokens $X$ and a global token $z_{\mathrm{cls}}$, we first project patch tokens to the codebook feature space
($Q=\mathrm{Proj}(X)$) and apply $\ell_2$ normalization to both $Q$ and $C$ to make similarity scores scale-stable.
We compute patch-to-concept similarity $S=\hat{Q}\hat{C}^\top/\tau$ and derive a per-patch evidence score by averaging the top-$r$ concept responses
(\textbf{Forensic Evidence Scoring}, FES). The top-$m$ patches by this score are selected as evidence patches.
Next, \textbf{Forensic Evidence Aggregator} (FEA) forms a global evidence vector $g$ by a softmax weighting over selected patches, 
where $\tau_w$ controls the sharpness of aggregation.
Finally, we output two logits: a standard classification head on $z_{\mathrm{cls}}$ and a concept head on $g$, and optimize them jointly with a weighted loss.
In our implementation, this design keeps the concept branch interpretable (explicit similarity to codebook entries) while remaining computationally light.

\begin{table}[t]
\centering
\setlength{\tabcolsep}{3pt}
\renewcommand{\arraystretch}{1.08}
\caption{\textbf{Cross-generator evaluation on GAN-family benchmarks (Accuracy, \%).}
Rows are grouped by training set: ProGAN (GAN family) vs.\ SDv1.4 (diffusion family).
\textbf{Bold} indicates the best result within the SDv1.4-trained group for each column.}
\label{tab:gan_family_acc}
\resizebox{\textwidth}{!}{%
\begin{tabular}{llcccccccc}
\toprule
Methods & Training Set & ProGAN & StyleGAN & BigGAN & CycleGAN & StarGAN & GauGAN & StyleGAN2 & Mean \\
\midrule
CNNSpot~\cite{wang2020cnn} & ProGAN & 100.0 & 90.2 & 71.2 & 87.6 & 94.6 & 81.4 & 86.9 & 87.4 \\
FreDect~\cite{frank2020leveraging} & ProGAN & 99.4 & 78.0 & 82.0 & 78.8 & 94.6 & 80.6 & 66.2 & 82.8 \\
Fusing~\cite{ju2022fusing} & ProGAN & 100.0 & 85.2 & 77.4 & 87.0 & 97.0 & 77.0 & 83.3 & 86.7 \\
LNP~\cite{liu2022detecting} & ProGAN & 99.7 & 91.8 & 77.8 & 84.1 & 100.0 & 75.4 & 94.6 & 89.1 \\
LGrad~\cite{tan2023learning} & ProGAN & 99.8 & 91.1 & 85.6 & 87.0 & 99.3 & 78.5 & 85.3 & 89.5 \\
UnivFD~\cite{ojha2023towards} & ProGAN & 99.8 & 84.9 & 95.1 & 98.3 & 95.8 & 99.5 & 75.0 & 92.6 \\
DIRE~\cite{wang2023dire} & ProGAN & 95.2 & 83.0 & 70.1 & 74.2 & 95.5 & 67.8 & 75.3 & 80.2 \\
PatchCraft~\cite{zhong2023patchcraft} & ProGAN & 100.0 & 92.8 & 95.8 & 70.2 & 100.0 & 71.6 & 89.6 & 88.6 \\
NPR~\cite{tan2024rethinking} & ProGAN & 99.8 & 97.7 & 84.4 & 96.1 & 99.4 & 82.5 & 98.4 & 94.0 \\
AIDE~\cite{yan2025sanity} & ProGAN & 100.0 & 99.6 & 84.0 & 98.5 & 99.9 & 73.3 & 98.0 & 93.3 \\
\midrule
DIRE~\cite{wang2023dire} & SDv1.4 & 52.8 & 51.3 & 49.7 & 49.6 & 46.7 & 51.2 & 51.7 & 50.4 \\
AIDE~\cite{yan2025sanity} & SDv1.4 & 69.3 & 71.1 & 66.9 & 74.4 & 80.2 & 60.4 & 72.5 & 70.7 \\
Effort~\cite{yan2025orthogonal} & SDv1.4 & 88.7 & 84.4 & 77.6 & \textbf{92.3} & \textbf{98.9} & 80.3 & 84.1 & 86.6 \\
\textbf{Ours} & SDv1.4 & \textbf{96.5} & \textbf{93.2} & \textbf{94.1} & 82.6 & 81.5 & \textbf{96.0} & \textbf{86.8} & \textbf{90.1} \\
\bottomrule
\end{tabular}%
}
\end{table}

\begin{table}[t]
\centering
\setlength{\tabcolsep}{3pt}
\renewcommand{\arraystretch}{1.08}
\caption{\textbf{Comparison on Chameleon.} Accuracy (\%) of different detectors (columns) for real/fake classification on the Chameleon dataset.}
\label{tab:chameleon_acc_wide}
\resizebox{\textwidth}{!}{%
\begin{tabular}{lccccccccccccc}
\toprule
Dataset & CNNSpot & FreDect & Fusing & GramNet & LNP & UnivFD & DIRE & PatchCraft & NPR & AIDE & Effort & AIGI-Holmes & Ours \\
\midrule
Chameleon & 60.1 & 56.9 & 57.1 & 61.0 & 55.6 & 55.6 & 59.7 & 56.3 & 58.1 & 62.6 & 63.2 & 75.9 & \textbf{84.4} \\
\bottomrule
\end{tabular}%
}
\end{table}

\section{Additional Results}
This section reports additional quantitative evaluations and one ablation study of our method 
(\cref{tab:mixed_generator_training,tab:recent_generators,tab:gan_family_acc,tab:chameleon_acc_wide,tab:cross_target_injection,tab:cgci_ablation_r_m_tauw}).

\paragraph{Mixed-generator training.}
To examine whether the gain depends on the SD~v1.4-only training setup, we further train DINOv3 on a GenImage-derived mixed-generator set containing both diffusion-style generators and BigGAN, following the experimental protocol of prior work \cite{zhou2025aigi}.
As shown in \cref{tab:mixed_generator_training}, the concept branch improves the mean accuracy from 96.72\% to 97.17\%.
This suggests that the benefit of forensic concepts is not tied to a codebook induced from a single diffusion-generator training distribution.

\begin{table}[t]
\centering
\small
\setlength{\tabcolsep}{4pt}
\renewcommand{\arraystretch}{1.08}
\caption{\textbf{Mixed-generator training results.}
Accuracy (\%) under a GenImage-derived mixed-generator training setting with both diffusion-style generators and BigGAN included in training.
The comparison between DINOv3 w/ and w/o concept shows whether the concept branch remains beneficial beyond the SD~v1.4-only setting.
\textbf{Bold} marks the best result per column.}
\label{tab:mixed_generator_training}
\begin{tabular}{lccccccccc}
\toprule
Methods & MJ & SD1.4 & SD1.5 & ADM & GLIDE & Wukong & VQDM & BigGAN & Mean \\
\midrule
DINOv3 w/o concept 
& 92.27 & 99.48 & 99.41 & 86.12 & 98.30 & \textbf{99.57} & \textbf{99.24} & \textbf{99.38} & 96.72 \\
DINOv3 w/ concept  
& \textbf{93.44} & \textbf{99.57} & \textbf{99.42} & \textbf{88.28} & \textbf{98.79} & 99.49 & 99.18 & 99.22 & \textbf{97.17} \\
\bottomrule
\end{tabular}
\vspace{-1em}
\end{table}

\paragraph{Evaluation on recent generators.}
To test whether the learned concepts remain effective beyond established public benchmarks, we construct a recent-generator test set based on \cite{li2026easier}.
The benchmark provides 4,320 fake images for each of 16 recent generators.
For each generator, we pair these fake images with 4,320 real images randomly sampled from GenImage, forming a balanced real/fake evaluation split.
As shown in \cref{tab:recent_generators}, our method achieves 96.23\% mean accuracy, substantially outperforming CLIP (67.32\%) and CLIP with codebook injection (90.85\%).
These gains indicate that our concept space captures transferable forensic cues beyond the original GenImage generators.

\begin{table}[t]
\centering
\small
\setlength{\tabcolsep}{4pt}
\renewcommand{\arraystretch}{1.08}
\caption{\textbf{Evaluation on recent generators.}
Accuracy (\%) on 16 recent image generators.
CLIP w/ inj. denotes CLIP with codebook injection.
\textbf{Bold} marks the best result per row.}
\label{tab:recent_generators}
\begin{tabular}{lccccccc}
\toprule
Generator & ResNet & EfficientNet & Swin-T & DeiT & CLIP & CLIP w/ inj. & Ours \\
\midrule
FLUX1 & 53.89 & 56.26 & 63.16 & 55.94 & 67.70 & 93.77 & \textbf{95.23} \\
FLUX2 & 57.65 & 64.47 & 64.68 & 63.41 & 63.84 & 90.41 & \textbf{94.84} \\
GPT-Image-1.5 & 54.28 & 52.81 & 55.67 & 57.87 & 59.05 & 89.91 & \textbf{95.75} \\
Gemini-2.0-Flash & 55.82 & 61.61 & 62.44 & 55.20 & 68.72 & 88.99 & \textbf{95.24} \\
HunyuanImage-3.0 & 52.29 & 54.80 & 54.33 & 56.02 & 55.53 & 90.94 & \textbf{99.17} \\
Infinity-8B & 56.99 & 58.84 & 64.91 & 68.29 & 84.18 & 96.27 & \textbf{99.63} \\
Nano-Banana-2 & 53.19 & 54.16 & 59.86 & 56.27 & 67.94 & 89.87 & \textbf{93.82} \\
Nano-Banana-Pro & 53.24 & 53.51 & 59.79 & 56.85 & 67.05 & 87.26 & \textbf{95.51} \\
SD-3.5-Large & 56.08 & 61.72 & 66.66 & 62.12 & 69.44 & 92.15 & \textbf{98.62} \\
Seedream-4 & 54.13 & 54.70 & 52.51 & 57.26 & 55.30 & 82.88 & \textbf{97.25} \\
Seedream-4.5 & 51.57 & 51.06 & 52.28 & 55.01 & 56.30 & 84.25 & \textbf{94.21} \\
Z-Image & 65.79 & 70.96 & 75.05 & 71.40 & 87.06 & 96.27 & \textbf{98.22} \\
Imagen-4 & 53.56 & 55.97 & 60.35 & 55.78 & 71.82 & \textbf{93.83} & 91.85 \\
Imagen-4-Ultra & 53.58 & 56.84 & 61.49 & 56.50 & 74.93 & \textbf{95.45} & 94.29 \\
Qwen-Image & 54.18 & 55.73 & 52.82 & 52.88 & 58.58 & 89.11 & \textbf{97.05} \\
Qwen-Image-2512 & 52.49 & 55.22 & 58.74 & 57.42 & 69.68 & 92.19 & \textbf{99.07} \\
\midrule
Mean & 54.92 & 57.42 & 60.30 & 58.64 & 67.32 & 90.85 & \textbf{96.23} \\
\bottomrule
\end{tabular}
\vspace{-1em}
\end{table}

\paragraph{Cross-generator generalization to GAN-family benchmarks.}
\cref{tab:gan_family_acc} evaluates whether a detector trained on a diffusion generator (SDv1.4) can transfer to GAN-family generators.
Within the SDv1.4-trained group, our method achieves the best mean accuracy (90.1\%), outperforming the strongest baseline in this group (Effort, 86.6\%).
Notably, we observe large gains on several GAN generators (e.g., ProGAN/StyleGAN/BigGAN) and on GauGAN, suggesting that generation-trace-derived concepts
can still capture transferable forensic cues beyond diffusion-only artifacts.
We also note that translation-style generators (CycleGAN/StarGAN) remain challenging under this training regime, indicating that 
evidence localization and concept selection could be further improved for image-to-image translation artifacts (see Limitations).

\paragraph{Chameleon benchmark.}
\cref{tab:chameleon_acc_wide} shows that our approach substantially improves robustness on the Chameleon benchmark, reaching 84.4\% accuracy, 
which exceeds prior detectors by a clear margin under the same evaluation protocol.
This result supports the motivation of introducing an explicit concept space and a generation-trace reference: 
when the test distribution becomes diverse and less ``template-like,'' relying on a single backbone's implicit feature space is often brittle, 
while concept-guided evidence aggregation provides an additional, structured decision pathway.

\paragraph{Cross-target codebook injection.}
\cref{tab:cross_target_injection} evaluates codebook injection after changing the target backbone to DeiT or ResNet while keeping the remaining setting unchanged.
CleanDIFT achieves the best result on both targets, improving DeiT from 69.73\% to 70.88\% and ResNet from 64.23\% to 66.26\%.
Among backbone-derived codebooks, DINO gives the strongest transfer on both DeiT and ResNet, consistent with its higher CKNNA alignment to diffusion traces.
These results suggest that the CKNNA--transfer relationship remains broadly informative across target backbones, although local deviations can still occur.

\paragraph{CGCI ablation.}
\cref{tab:cgci_ablation_r_m_tauw} ablates CGCI-specific hyperparameters in the concept-guided branch.
Performance is most sensitive to the temperature terms: a larger similarity temperature $\tau$ in \cref{eq:similarity} and overly smooth aggregation via $\tau_w$ in FEA (\cref{eq:fea}) both degrade accuracy, suggesting that CGCI benefits from relatively sharp patch--concept responses and moderately peaked evidence weights.
In contrast, the discrete controls in CGCI (FES top-$r$ pooling and the TopM budget $m$; \cref{eq:score,eq:select_topm}) are comparatively stable within reasonable ranges, while extreme settings tend to introduce noisy concepts and dilute salient cues.

\begin{table}[t]
\centering
\small
\setlength{\tabcolsep}{4pt}
\renewcommand{\arraystretch}{1.08}
\caption{\textbf{Cross-target codebook injection results.}
Accuracy (\%) under different target backbones and codebook sources.
The ``w/o inj.'' column denotes the target backbone without codebook injection.
\textbf{Bold} marks the best result per row.}
\label{tab:cross_target_injection}
\begin{tabular}{lcccccccc}
\toprule
Target & w/o inj. & DINO & DeiT & CLIP & Swin-T & EffNet & ResNet & CleanDIFT \\
\midrule
DeiT   
& 69.73 & 70.73 & 70.58 & 70.39 & 70.32 & 70.62 & 70.20 & \textbf{70.88} \\
ResNet 
& 64.23 & 65.65 & 65.32 & 65.15 & 65.13 & 64.96 & 64.82 & \textbf{66.26} \\
\bottomrule
\end{tabular}
\vspace{-1em}
\end{table}

\begin{table}[t]
\centering
\small
\setlength{\tabcolsep}{4pt}
\renewcommand{\arraystretch}{1.08}
\caption{\textbf{CGCI hyperparameter ablation.}
Mean accuracy (ACC, \%) on GenImage with CLIP, trained on the SDv1.4 training split and evaluated on the official GenImage test splits using the same diffusion-derived codebook.
$\Delta$ denotes the absolute ACC change relative to the default setting
($r{=}8, m{=}20, \tau{=}0.1, \tau_w{=}0.05$).}
\label{tab:cgci_ablation_r_m_tauw}
\begin{tabular}{lccc@{\hspace{1.2em}}lccc}
\toprule
Setting & Value & ACC & $\Delta$ &
Setting & Value & ACC & $\Delta$ \\
\midrule
\multicolumn{4}{l}{\textit{Similarity temperature $\tau$}} &
\multicolumn{4}{l}{\textit{TopM evidence budget $m$}} \\
$\tau$ & 0.2 & 83.8 & -4.4 &
$m$    & 10  & 87.4 & -0.8 \\
$\tau$ & 0.5 & 84.1 & -4.1 &
$m$    & 15  & 86.7 & -1.5 \\
$\tau$ & 1.0 & 84.2 & -4.0 &
$m$    & 25  & 87.2 & -1.0 \\
       &     &      &      &
$m$    & 30  & 86.4 & -1.8 \\
\addlinespace[0.25em]

\multicolumn{4}{l}{\textit{FES top-$r$ pooling}} &
\multicolumn{4}{l}{\textit{FEA aggregation temperature $\tau_w$}} \\
$r$ & 1 & 87.1 & -1.1 &
$\tau_w$ & 0.04 & 85.4 & -2.8 \\
$r$ & 2 & 85.8 & -2.4 &
$\tau_w$ & 0.06 & 85.7 & -2.5 \\
$r$ & 7 & 86.4 & -1.8 &
$\tau_w$ & 0.50 & 84.2 & -4.0 \\
$r$ & 9 & 85.9 & -2.3 &
$\tau_w$ & 1.00 & 84.2 & -4.0 \\
\midrule
\multicolumn{6}{l}{\textbf{Default:} $r{=}8, m{=}20, \tau{=}0.1, \tau_w{=}0.05$} &
\textbf{88.2} & \textbf{0.0} \\
\bottomrule
\end{tabular}
\vspace{-1em}
\end{table}

\section{Codebook Center Visualizations}
\label{sec:vis_supp}
\begin{figure}[t]
\centering
\includegraphics[width=\columnwidth]{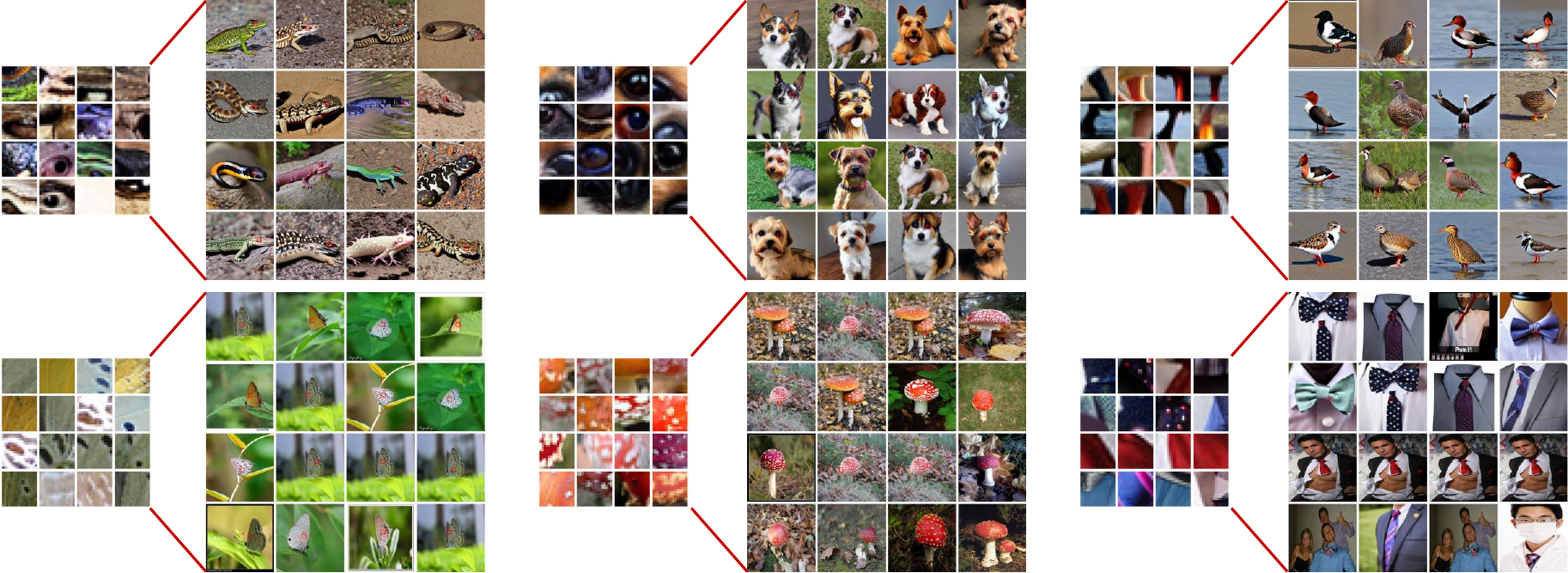}
\caption{\textbf{Stable Diffusion 1.4, DINOv3.}}
\label{fig:codebook_vis_sd14_dinov3}
\end{figure}

\begin{figure}[t]
\centering
\includegraphics[width=\columnwidth]{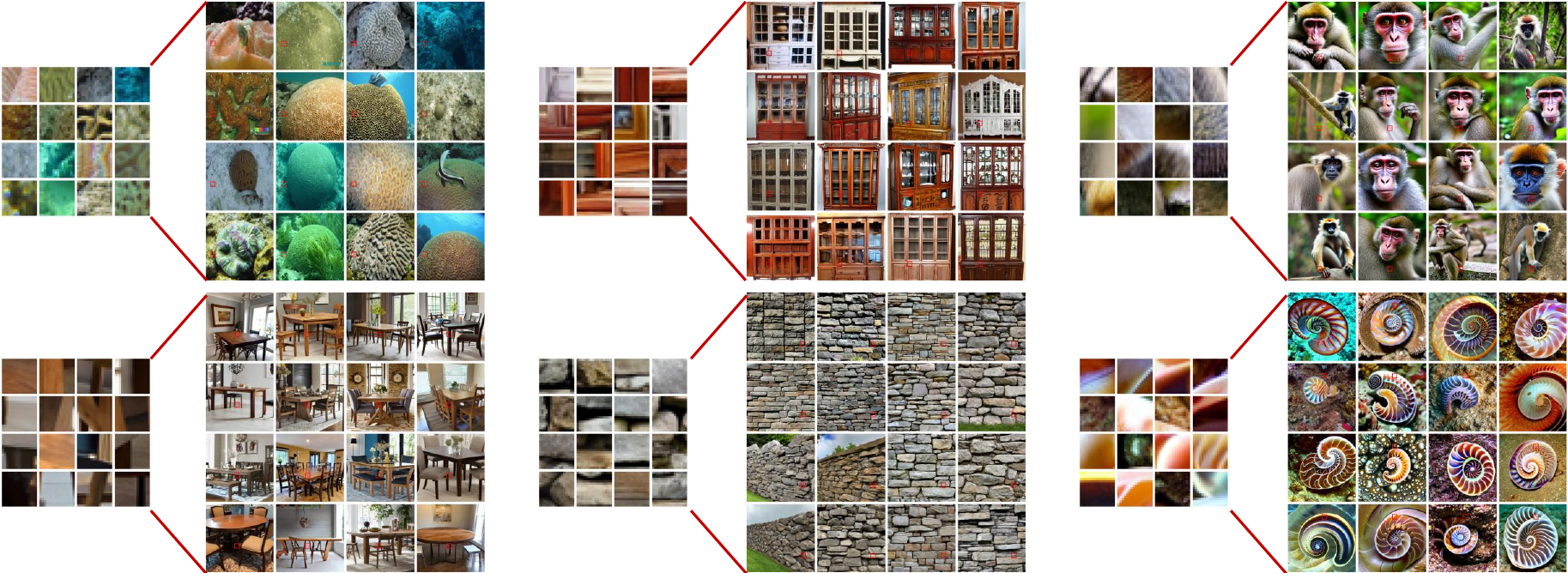}
\caption{\textbf{Stable Diffusion 1.4, DeiT.}}
\label{fig:codebook_vis_sd14_deit}
\end{figure}

\begin{figure}[H]
\centering
\includegraphics[width=\columnwidth]{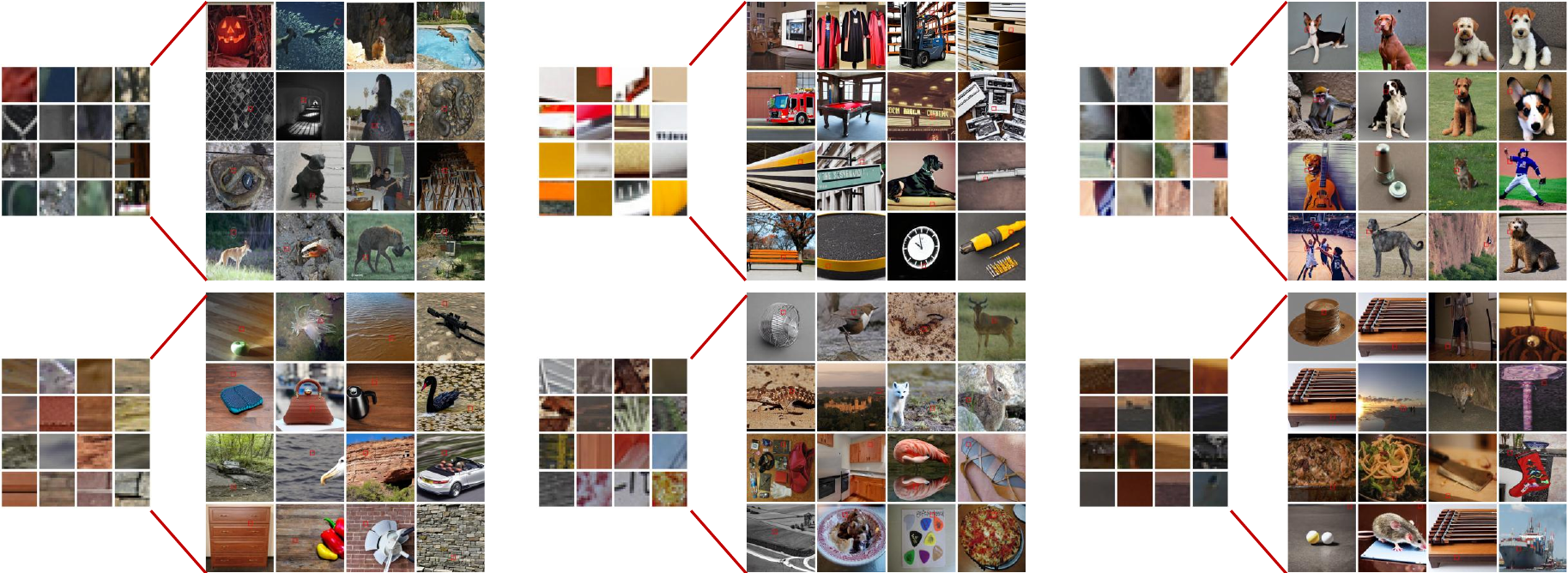}
\caption{\textbf{Stable Diffusion 1.4, ResNet.}}
\label{fig:codebook_vis_sd14_resnet}
\end{figure}

\begin{figure}[t]
\centering
\includegraphics[width=\columnwidth]{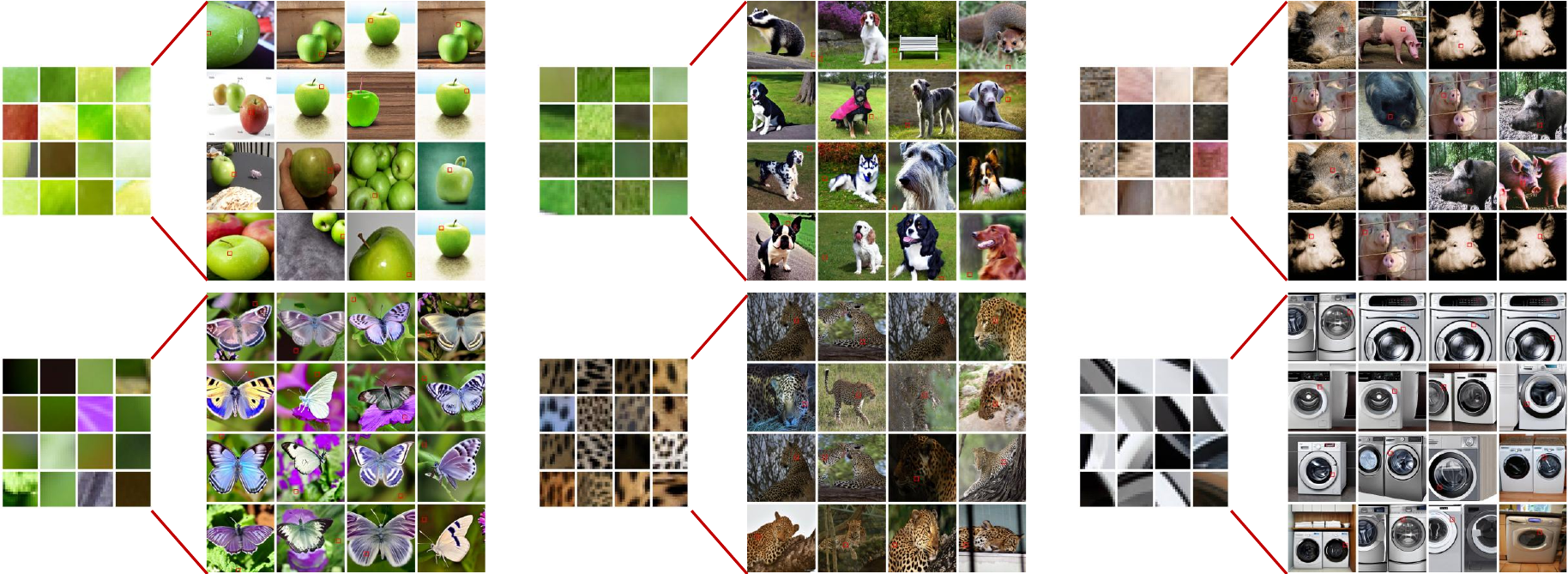}
\caption{\textbf{Stable Diffusion 1.4, CLIP.}}
\label{fig:codebook_vis_sd14_clip}
\end{figure}

\begin{figure}[H]
\centering
\includegraphics[width=\columnwidth]{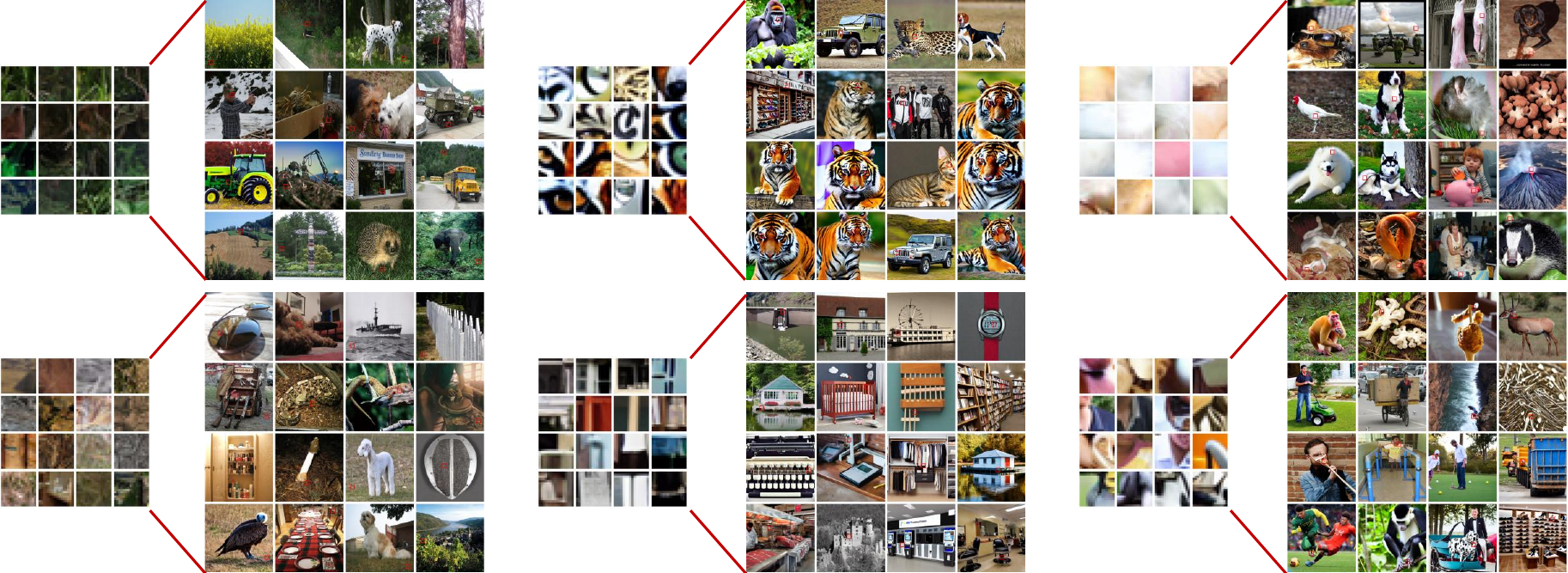}
\caption{\textbf{Stable Diffusion 1.4, EfficientNet.}}
\label{fig:codebook_vis_sd14_efficientnet}
\end{figure}

\begin{figure}[H]
\centering
\includegraphics[width=\columnwidth]{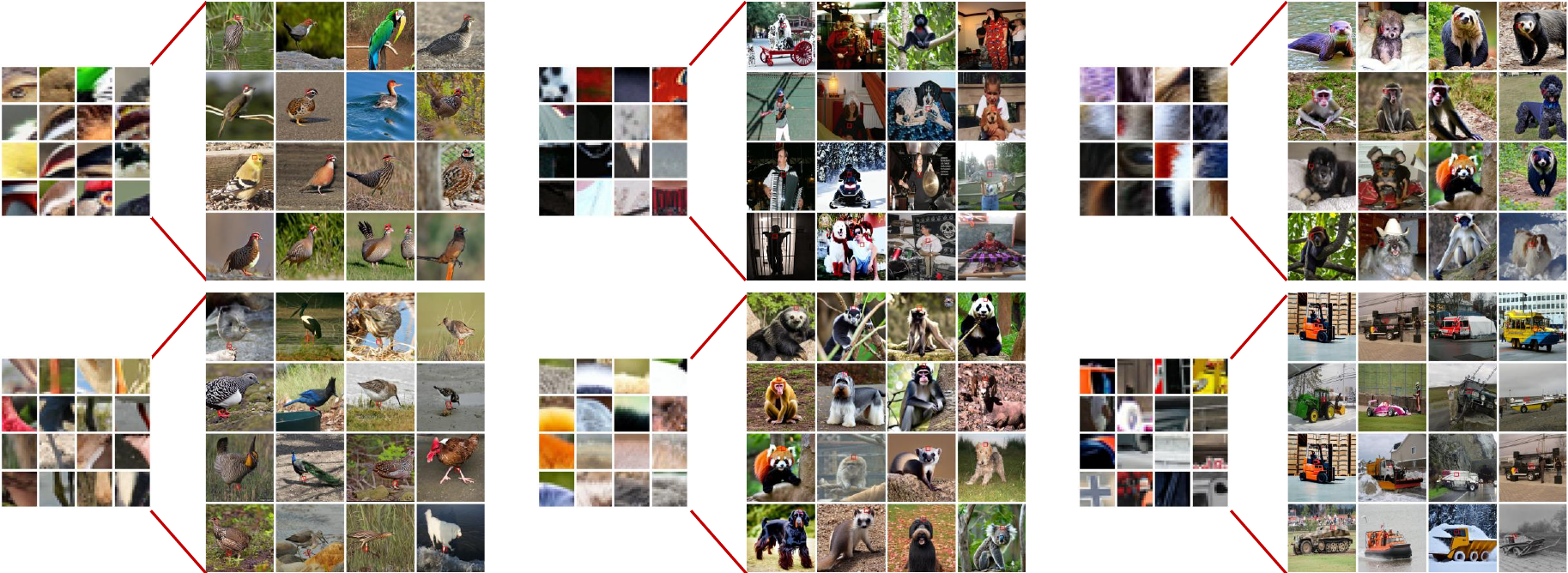}
\caption{\textbf{Stable Diffusion 1.4, CleanDIFT.}}
\label{fig:codebook_vis_sd14_cleandift}
\end{figure}

\paragraph{Backbone-specific codebooks on SDv1.4 (linked to CKNNA).}
\cref{fig:codebook_vis_sd14_dinov3}--\cref{fig:codebook_vis_sd14_cleandift} visualize evidence-aligned codebooks extracted under different backbones
(DINOv3, DeiT, CLIP, Swin-T, EfficientNet \cite{tan2019efficientnet}, ResNet) and the generation-trace reference (CleanDIFT) on SDv1.4.
These qualitative patterns are consistent with the quantitative backbone--trace alignment measured by CKNNA:
backbones with higher CKNNA tend to yield codebook entries that are (i) more visually cohesive around the cluster centers, and
(ii) more stable in where they are re-identified across images, suggesting that their induced concepts reflect more repeatable forensic structure.
In particular, DINOv3 exhibits the clearest and most repeatable prototypes, matching its strong alignment to diffusion traces,
while weaker-aligned backbones more often produce heterogeneous centers or less consistent matches, indicating reduced concept stability.

\begin{figure}[H]
\centering
\includegraphics[width=\columnwidth]{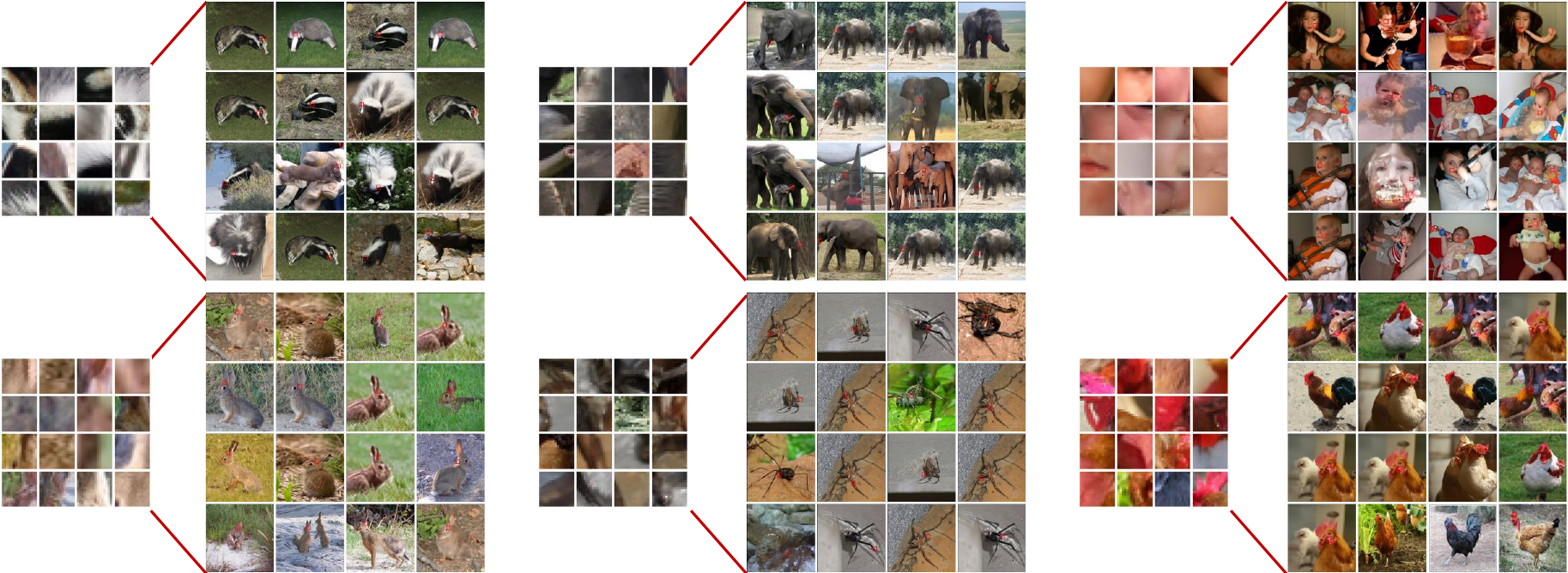}
\caption{\textbf{ADM, DINOv3.}}
\label{fig:codebook_vis_adm_dino}
\end{figure}

\begin{figure}[H]
\centering
\includegraphics[width=\columnwidth]{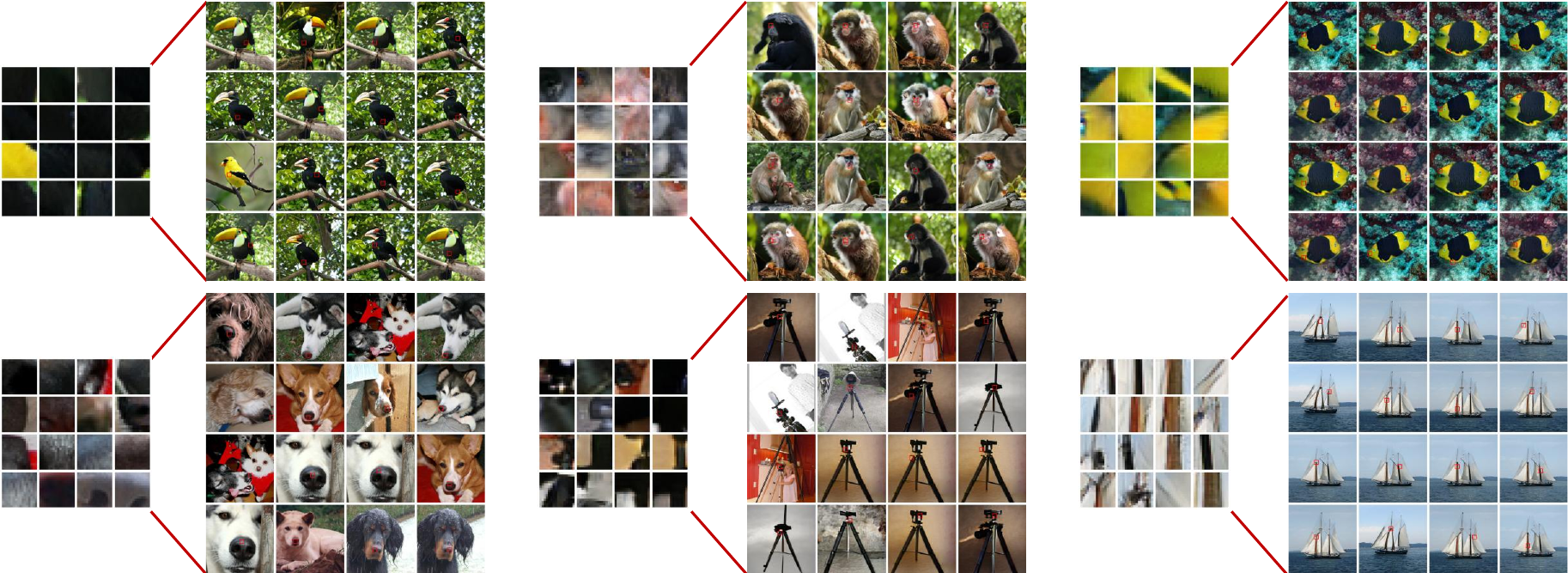}
\caption{\textbf{BigGAN, DINOv3.}}
\label{fig:codebook_vis_biggan_dino}
\end{figure}

\begin{figure}[H]
\centering
\includegraphics[width=\columnwidth]{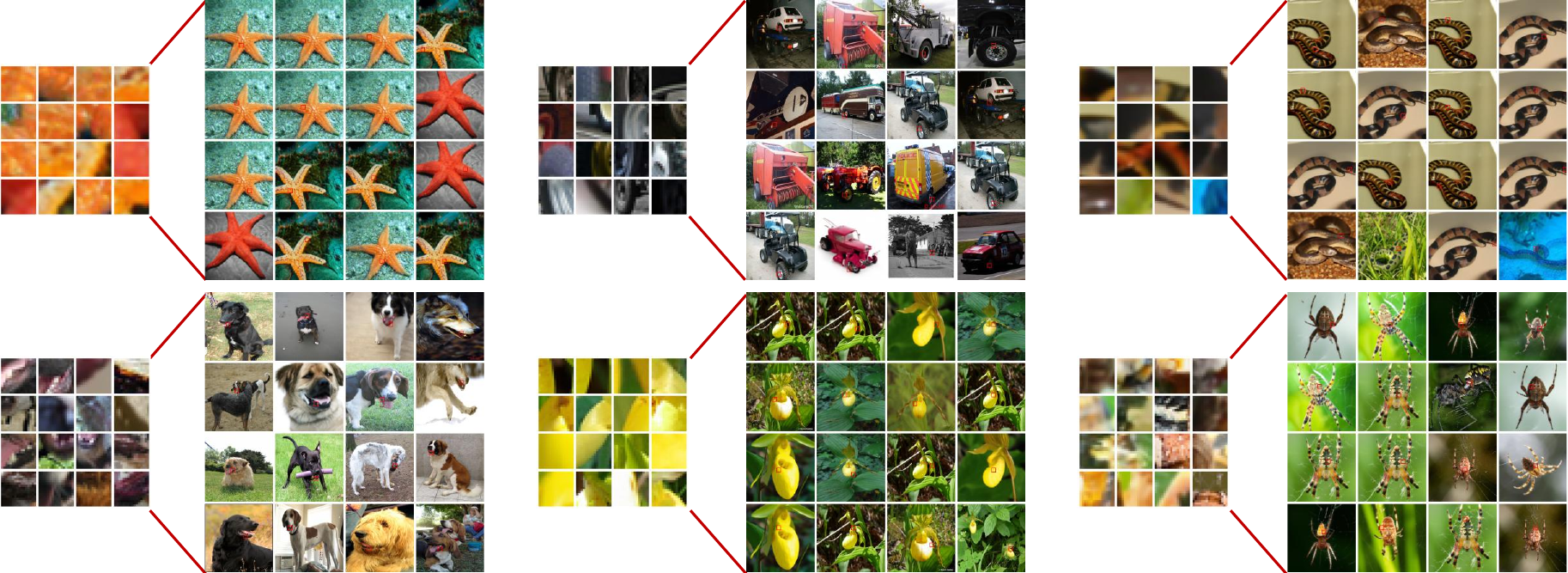}
\caption{\textbf{GLIDE, DINOv3.}}
\label{fig:codebook_vis_glide_dino}
\end{figure}

\begin{figure}[H]
\centering
\includegraphics[width=\columnwidth]{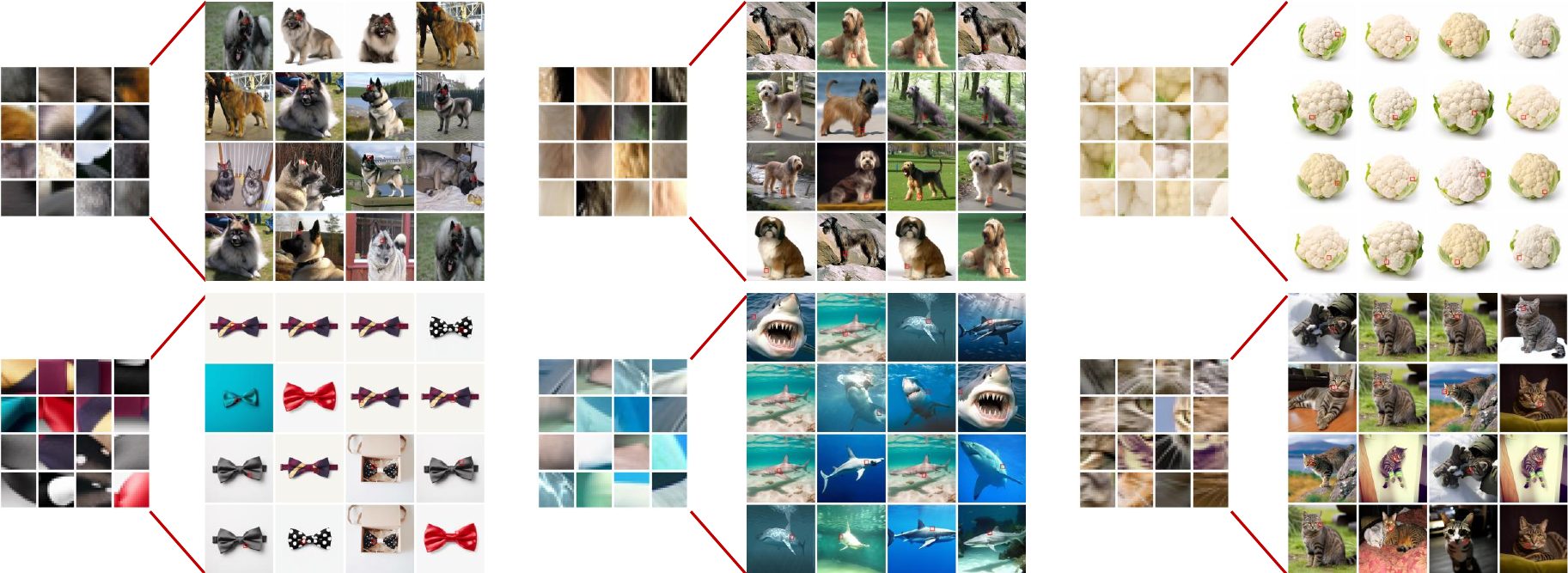}
\caption{\textbf{Midjourney, DINOv3.}}
\label{fig:codebook_vis_midjourney_dino}
\end{figure}

\begin{figure}[H]
\centering
\includegraphics[width=\columnwidth]{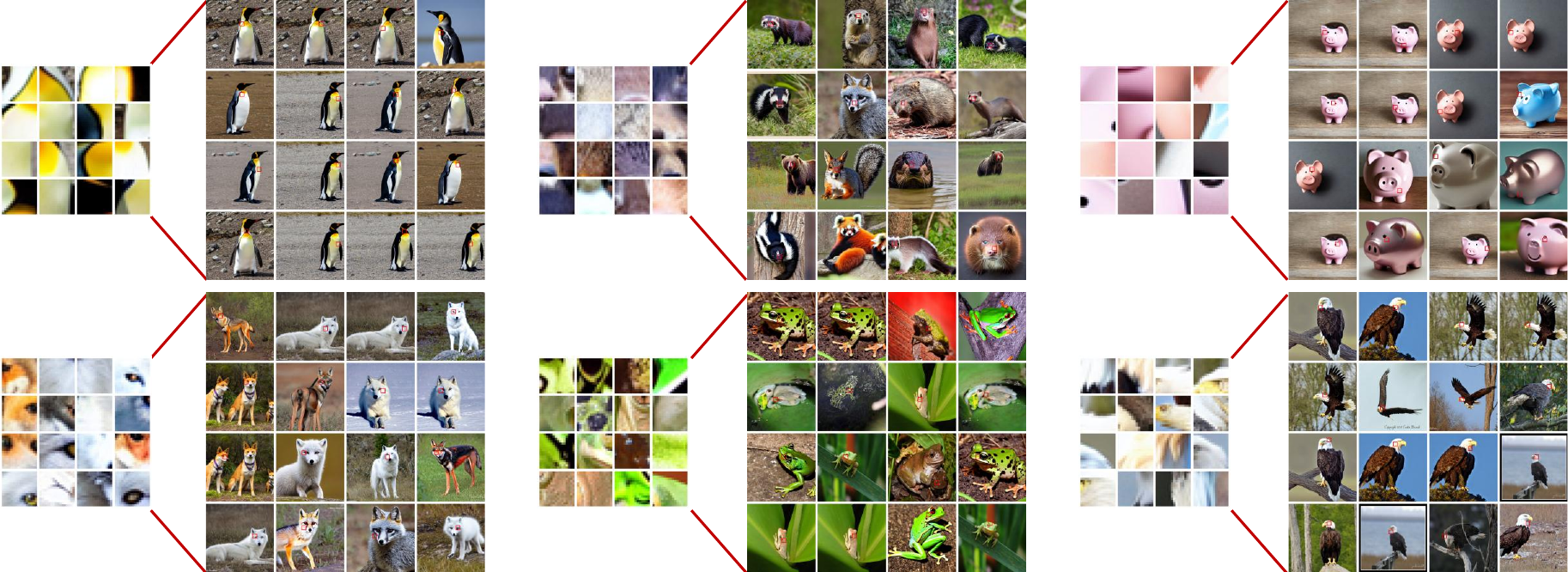}
\caption{\textbf{Stable Diffusion 1.5, DINOv3.}}
\label{fig:codebook_vis_sd15_dino}
\end{figure}

\begin{figure}[H]
\centering
\includegraphics[width=\columnwidth]{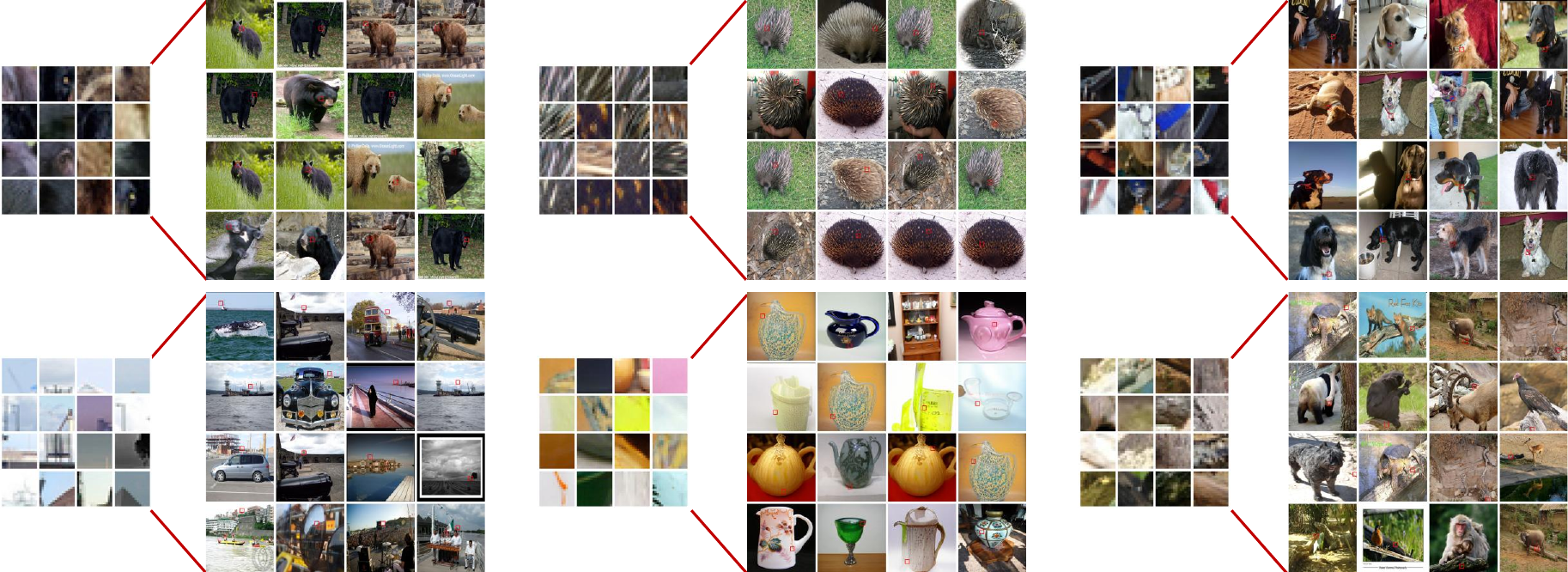}
\caption{\textbf{VQDM, DINOv3.}}
\label{fig:codebook_vis_vqdm_dino}
\end{figure}

\begin{figure}[H]
\centering
\includegraphics[width=\columnwidth]{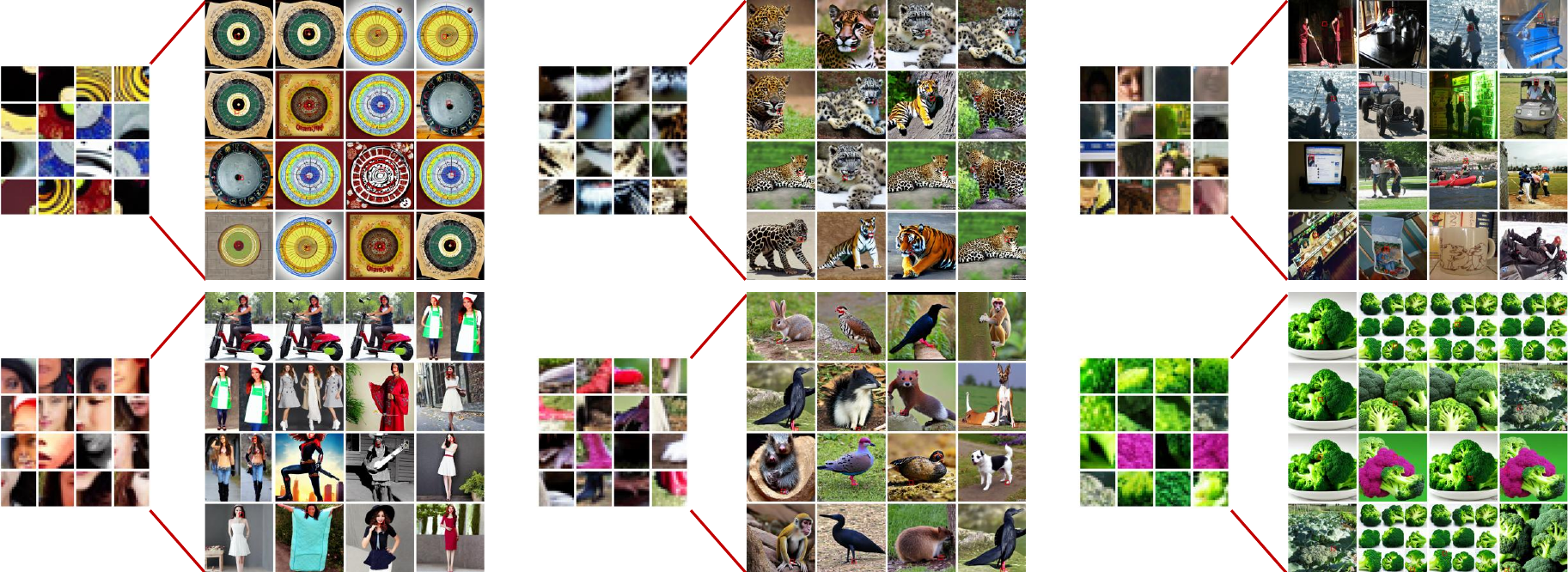}
\caption{\textbf{Wukong, DINOv3.}}
\label{fig:codebook_vis_wukong_dino}
\end{figure}

\begin{figure}[H]
\centering
\includegraphics[width=\columnwidth]{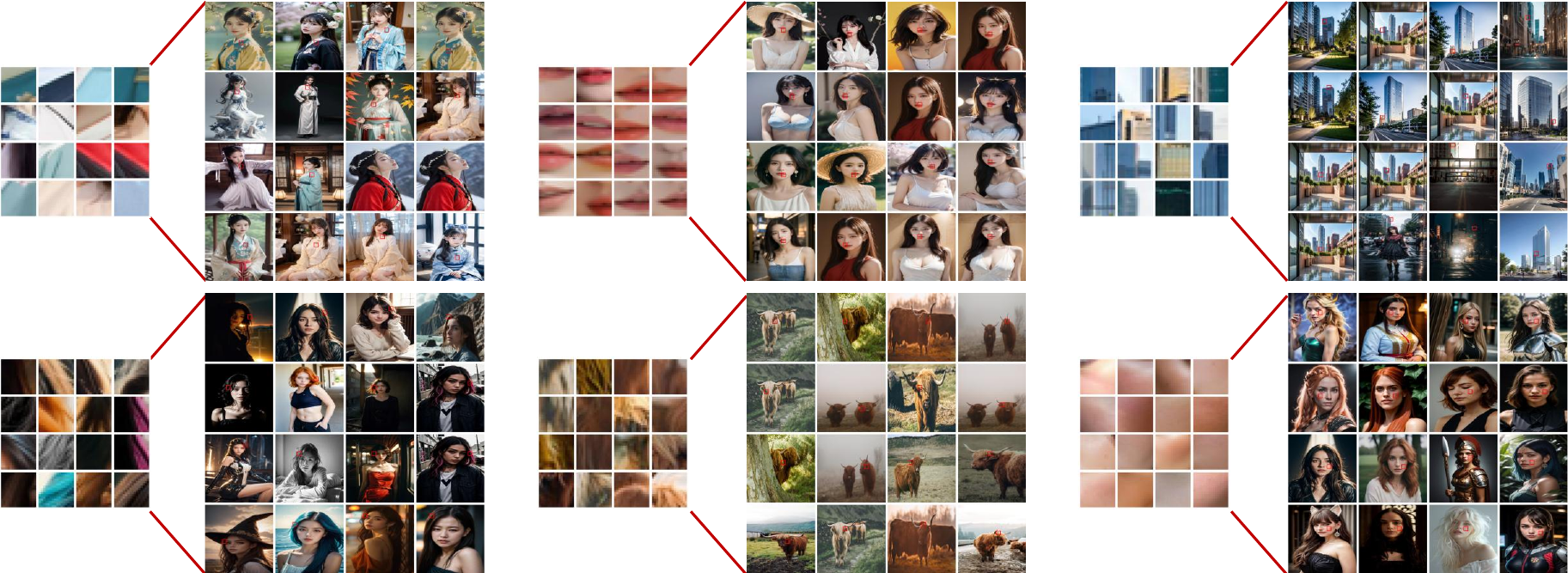}
\caption{\textbf{Chameleon, DINOv3.}}
\label{fig:codebook_vis_chameleon_dino}
\end{figure}

\begin{figure}[H]
\centering
\includegraphics[width=\columnwidth]{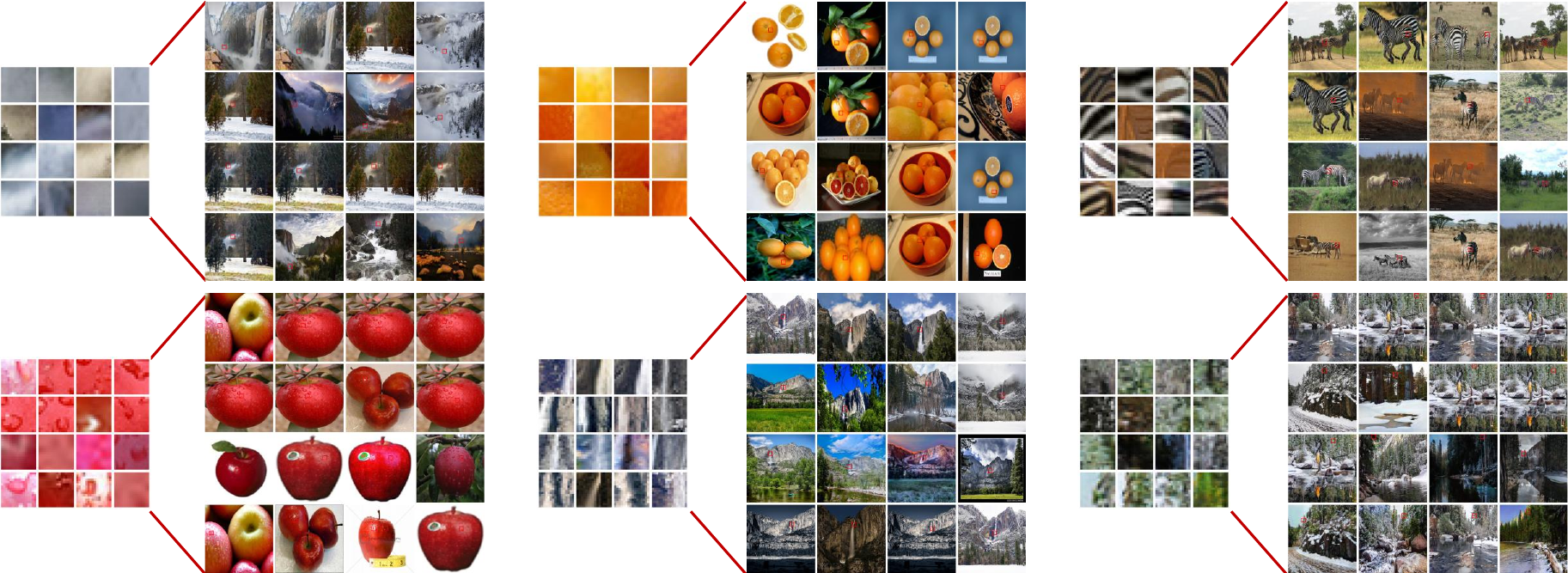}
\caption{\textbf{CycleGAN, DINOv3.}}
\label{fig:codebook_vis_cyclegan_dino}
\end{figure}

\begin{figure}[H]
\centering
\includegraphics[width=\columnwidth]{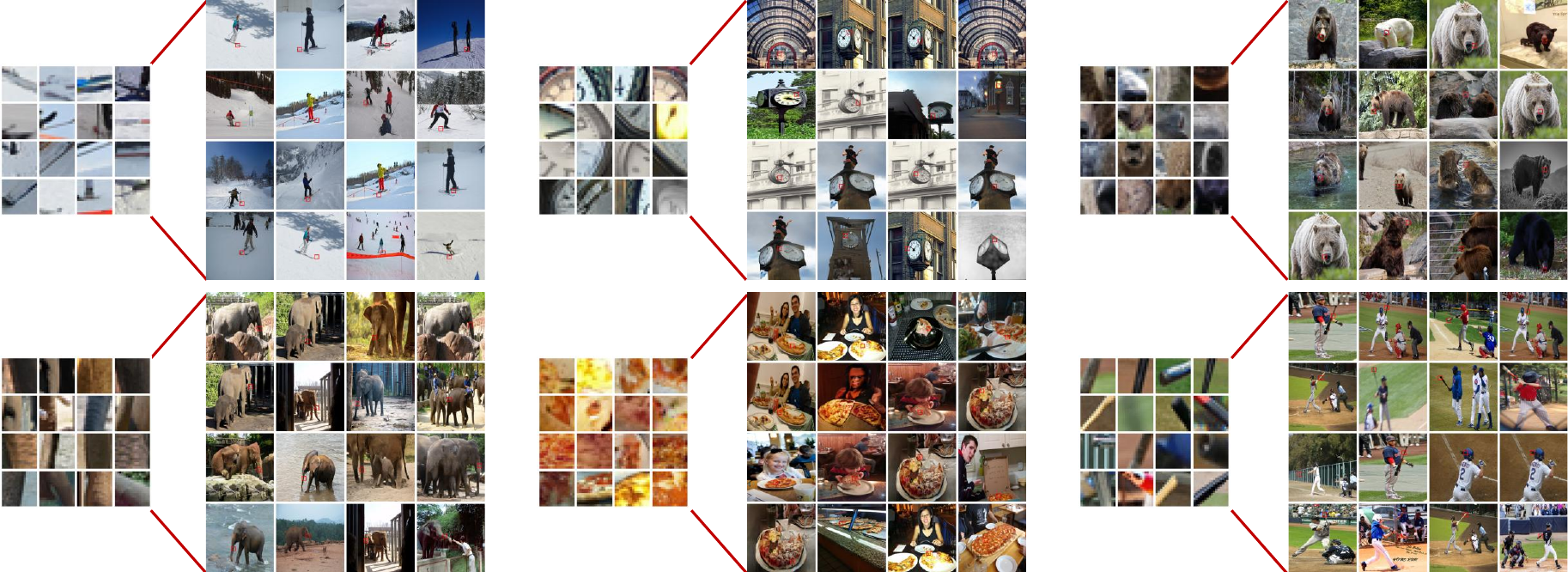}
\caption{\textbf{GauGAN, DINOv3.}}
\label{fig:codebook_vis_gaugan_dino}
\end{figure}

\begin{figure}[H]
\centering
\includegraphics[width=\columnwidth]{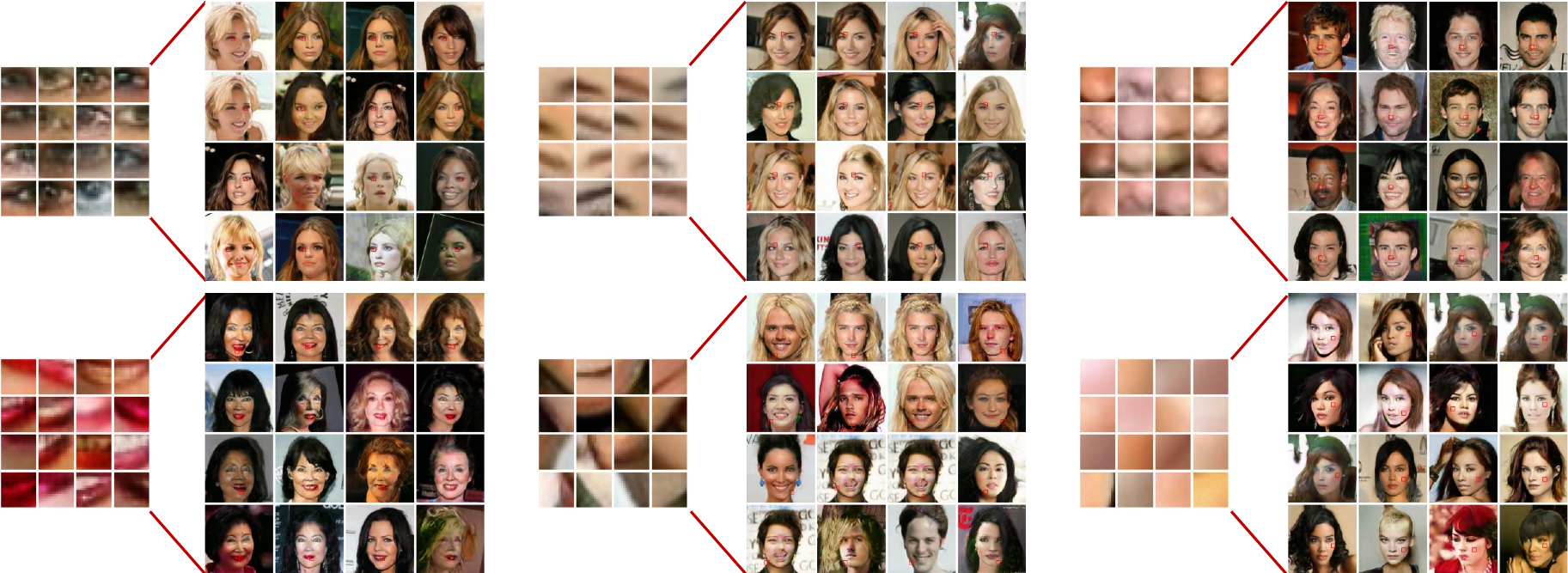}
\caption{\textbf{StarGAN, DINOv3.}}
\label{fig:codebook_vis_stargan_dino}
\end{figure}

\begin{figure}[H]
\centering
\includegraphics[width=\columnwidth]{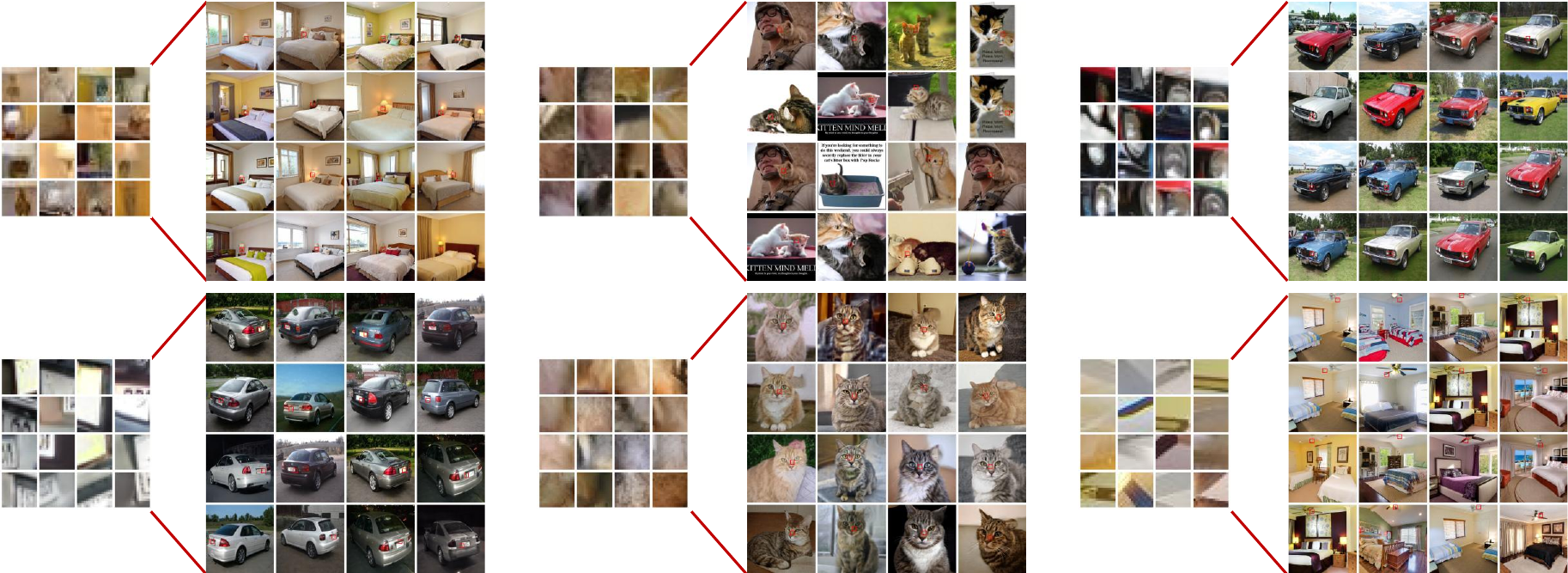}
\caption{\textbf{StyleGAN, DINOv3.}}
\label{fig:codebook_vis_stylegan_dino}
\end{figure}

\paragraph{Generator-specific codebooks under DINOv3 (why DINO appears most stable).}
\cref{fig:codebook_vis_adm_dino}--\cref{fig:codebook_vis_stylegan_dino} present DINOv3 codebook centers across diverse generators.
Many prototypes correspond to coherent micro-textures and low-level structures, and their matched locations span multiple semantic regions,
indicating that the clusters capture generator-related forensic cues rather than object semantics.
This behavior aligns with the intent of CKNNA: if a backbone’s evidence neighborhood structure is consistent with diffusion traces,
its induced concepts should remain stable under content variation yet sensitive to generation artifacts.
The strong visual regularity of DINOv3 concepts across generators provides additional human-verifiable support for its higher CKNNA scores.

\section{Limitations and Future Work}
A practical limitation of our study is the limited quality and coverage of current public benchmarks for AIGI forensics.
Across datasets, real/fake labels are often coarse, generator metadata can be incomplete, and the space of post-processing pipelines
(e.g., resizing, compression, tone mapping, re-encoding, and platform-specific transforms) is not systematically controlled.
These factors introduce unavoidable confounders when measuring transferability and can obscure the full potential of concept-based evidence,
especially when the objective is to isolate stable generation traces from incidental dataset biases.

A promising direction is to develop or adopt higher-quality, more diagnostic benchmarks that offer finer-grained generator annotations,
controlled post-processing factors, and broader coverage of both synthesis and translation pipelines.
Such datasets would enable more rigorous evaluation of (i) which forensic concepts are truly generator-invariant, (ii) how alignment metrics such as CKNNA
respond to controlled shifts, and (iii) how codebook injection behaves under realistic deployment transformations.
We also expect that richer annotations (e.g., source generator family, editing operations, and rendering pipelines) can support more structured learning settings,
including multi-branch or continual extensions of concept spaces, while preserving the interpretability benefits of our current framework.

\paragraph{Ethics \& Reproducibility.}
Our experiments use publicly available benchmarks (e.g., ImageNet-based real images and synthetic images generated by widely used generators) with appropriate citations.
The proposed detector can support benign applications such as content provenance and integrity verification, but it could also be misused for surveillance or censorship; we therefore encourage responsible deployment and transparent reporting of failure cases under distribution shift.
To facilitate reproducibility, we will release code, trained checkpoints, evaluation scripts, and preprocessing details.

\end{document}